\documentclass[runningheads]{llncs}

\usepackage{eccv}

\usepackage{eccvabbrv}

\usepackage{graphicx}
\usepackage{booktabs}

\usepackage[accsupp]{axessibility}  

\usepackage{hyperref}

\usepackage{orcidlink}
\usepackage[dvipsnames]{xcolor}

\usepackage{tikz}
\usepackage{amsmath}
\usepackage{pgfplots}
\usepackage{pgfplotstable}
\usetikzlibrary{arrows.meta}
\pgfplotsset{compat=newest}
\usepackage{dirtytalk}
\usepackage{nicefrac} 
\usepackage{colortbl}

\usepackage{multirow}
\usepackage[normalem]{ulem}
\usepackage{amssymb} 
\usepackage{appendix}
\usepackage{amsmath}
\usepackage{adjustbox}
\usepackage{enumitem}
\usepackage{microtype}

\usepackage{algorithm}
\usepackage{algpseudocodex}

\definecolor{C0}{HTML}{1f77b4}
\definecolor{C1}{HTML}{ff7f0e}

\newcommand{\abs}[1]{\lvert #1 \rvert}
\newcommand{\order}[1]{\mathcal{O}(#1)}

\makeatletter
\newcommand\@chapapp{Appendix}
\makeatother

\makeatother

\begin{document}

\title{Instance-dependent Noisy-label Learning with Graphical Model Based Noise-rate Estimation} 

\author{Arpit Garg\inst{1} \and
Cuong Nguyen\inst{3} \and
Rafael Felix\inst{1} \and Thanh-Toan Do\inst{2} \and Gustavo Carneiro\inst{3}}


\institute{Australian Institute for Machine Learning, University of Adelaide, Australia \and Department of Data Science and AI, Monash University, Australia \and Centre for Vision, Speech and Signal Processing, University of Surrey, UK
}

\maketitle

\begin{abstract} 
Deep learning faces a formidable challenge when handling noisy labels, as models tend to overfit samples affected by label noise. This challenge is further compounded by the presence of instance-dependent noise (IDN), a realistic form of label noise arising from ambiguous sample information. To address IDN, Label Noise Learning (LNL) incorporates a sample selection stage to differentiate clean and noisy-label samples. This stage uses an arbitrary criterion and a pre-defined curriculum that initially selects most samples as noisy and gradually decreases this selection rate during training. Such curriculum is sub-optimal since it does not consider the actual label noise rate in the training set. This paper addresses this issue with a new noise-rate estimation method that is easily integrated with most state-of-the-art (SOTA) LNL methods to produce a more effective curriculum. Synthetic and real-world benchmarks’ results demonstrate that integrating our approach with SOTA LNL methods improves accuracy in most cases.\footnote{\scriptsize Code is available at \url{https://github.com/arpit2412/NoiseRateLearning}. \newline Supported by the Engineering and Physical Sciences Research Council (EPSRC) through grant EP/Y018036/1 and the Australian Research Council (ARC) through grant FT190100525.} 
\end{abstract}

\section{Introduction}\label{sec:introduction}

Deep neural networks (DNNs) have demonstrated their effectiveness in various domains, including vision~\cite{yang2022unsupervised}, language~\cite{trummer2022bert}, and medicine~\cite{shamshad2023transformers}. However, its efficacy is largely dependent on high-quality training data, which can be resource-intensive to obtain~\cite{tu2022debtfree}.
Although cost-effective labelling techniques, such as data mining~\cite{feng2021ssr} and crowdsourcing~\cite{song2022learning}, offer cheaper alternatives, they often compromise label quality~\cite{song2022learning}. Consequently, this can lead to erroneous labels in real-world datasets~\cite{kim2021fine}. This is a relevant issue because even slight inaccuracies in the labels can significantly affect the performance of DNN due to their inherent memorisation capabilities~\cite{zhang2021understanding, neyshabur2017exploring}. This has prompted the development of algorithms resilient to noisy-label learning, with the aim of training models despite inaccuracies in training data labels. 
Although various strategies exist, our work uniquely proposes an approach centered on the estimation of the label noise rate from the training data, which is notably absent from current methodologies.

There are two types of label noise, namely instance-independent noise (IIN)~\cite{han2018co} and instance-dependent noise (IDN)~\cite{xia2020part}.
Such type of label noise typically influences the design principles of noisy-label learning algorithms. 
For example, IIN focuses on mislabellings that are independent of sample information~\cite{han2018co}, where estimating the underlying label transition matrix is a common way of handling this type of noise~\cite{yao2020dual}. 
On the other hand, in the more realistic IDN, mislabelling is due to both sample information and true class labels~\cite{xia2020part}, which generally require the combination of many label noise learning techniques, such as robust loss functions~\cite{zhang2018generalized, liu2020early}, and noisy-label sample selection~\cite{li2020dividemix, zhao2022centrality}. 
Of the strategies mentioned above, sample selection approaches that classify training data into clean and noisy samples have produced competitive results on many benchmarks~\cite{li2020dividemix, cordeiro2021propmix, feng2021ssr, kim2021fine, garg2023instance}. 
Such sample selection techniques require the definition of a classification criterion and a selection curriculum.
Many studies on this topic focus on developing new sample selection criteria, such as the small-loss hypothesis~\cite{li2020dividemix}, which states that noisy-label samples have higher loss values than clean-label samples, particularly at the early stage of training~\cite{arpit2017closer}. 
Another criterion type is the feature-based one.
An example of such criterion is FINE~\cite{kim2021fine} that discriminates clean and noisy-label samples via the distance to class-specific eigenvectors. 
In this technique, clean-label samples tend to lie closer to the class-specific dominant eigenvector of the latent representations than noisy-label samples.
Another type of criterion is proposed in SSR~\cite{feng2021ssr}, which introduces a selection criterion based on the K-nearest-neighbor (KNN) classification in the feature space. Furthermore, CC~\cite{zhao2022centrality} uses a two-stage sampling procedure, including class-level feature clustering followed by a consistency score.  
An equally important problem in sample selection is the definition of the curriculum for selecting clean training samples, but it has received comparatively less attention.

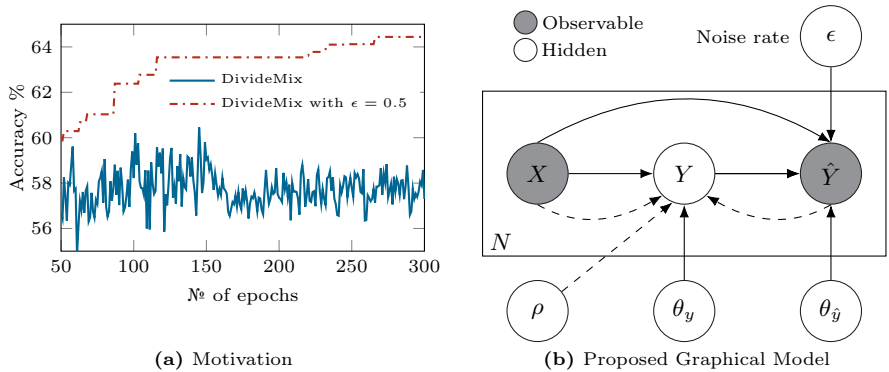
\begin{figure*}[t!]
    \centering
    \begin{subfigure}{0.495\linewidth}
        \centering
        \begin{tikzpicture}
    \pgfplotstableread[col sep=comma, header=true]{csvs/motivation.tex} \myTable
    \begin{axis}[
        height = 0.5\linewidth,
        width = 0.8\linewidth,
        xlabel={\textnumero~of epochs},
        xlabel style={font=\scriptsize},
        xticklabel style = {font=\scriptsize},
        xmin=50,
        xmax=300,
        xtick={50, 100, 150, 200, 250, 300},
        ylabel={Accuracy \%},
        ylabel style={font=\scriptsize, yshift=-0.5em},
        yticklabel style = {font=\scriptsize},
        ymin=55,
        ymax=65,
        legend entries={DivideMix, DivideMix with \(\epsilon = 0.5\)},
        legend style={draw=none, font=\tiny, at={(0.99,0.7)}, anchor=east},
        legend image post style={scale=1.},
        legend cell align={left},
        at={(0.98,0.5)}, 
        anchor=east, 
        scale only axis
    ]
    \addplot[mark=none, MidnightBlue, thick, solid] table[x={epochs}, y={DivideMix}]{\myTable};
    \addplot[mark=none, BrickRed, solid, thick, dashdotted] table[x={epochs}, y={fixE}]{\myTable};
    \end{axis}
\end{tikzpicture}
        \caption{Motivation}
        \label{fig:motivation_graph}
    \end{subfigure}
    \hfill
    \begin{subfigure}{0.495\linewidth}
        \centering
        \begin{tikzpicture}
	\pgfmathsetmacro{\yshift}{4.5}
	\pgfmathsetmacro{\xshift}{6}
	\pgfmathsetmacro{\minsize}{2.5}
	
	\node[circle, draw=Black, fill=Gray, minimum size=\minsize em, inner sep=0em] at (0, 0) (x) {\(X\)};
	
	\node[circle, draw=Black, minimum size=\minsize em, inner sep=0em] at ([xshift=\xshift em, yshift=0em]x) (y) {\(Y\)};
	
	\node[circle, draw=Black, fill=Gray, minimum size=\minsize em, inner sep=0em] at ([xshift=\xshift em, yshift=0em]y) (yhat) {\(\hat{Y}\)};

	\node[circle, draw=Black, minimum size=\minsize em, inner sep=0em] at ([xshift=0em, yshift=-1.25*\yshift em]y) (theta_y) {\(\theta_{y}\)};
	
	\node[circle, draw=Black, minimum size=\minsize em, inner sep=0em] at ([xshift=0em, yshift=-1.25*\yshift em]yhat) (theta_yhat) {\(\theta_{\hat{y}}\)};
	
	\node[circle, draw=Black, minimum size=\minsize em, inner sep=0em] at ([xshift=0em, yshift=1.25*\yshift em]yhat) (noise) {\(\epsilon\)};
        \node[] at ([xshift=-0.6*\xshift em]noise) {\scriptsize Noise rate};
	
	\node[circle, draw=Black, minimum size=\minsize em, inner sep=0em] at ([yshift=-1.25*\yshift em]x) (rho) {\(\rho\)};
	
	\node[circle, draw=Black, fill=Gray, minimum size=0.375em, anchor=east] at ([yshift=1.375*\yshift em]x) (observable) {};
	\node[] at ([xshift=2.75em]observable) { \scriptsize Observable};
	
	\node[circle, draw=Black, minimum size=0.375em, anchor=east] at ([yshift=1.125*\yshift em]x) (hidden) {};
	\node[] at ([xshift=2em]hidden) {\scriptsize Hidden};

	\draw[] ([xshift=-0.375 * \xshift em, yshift=-0.75 * \yshift em]x) rectangle ([xshift=0.375 * \xshift em, yshift=0.75 * \yshift em]yhat);
	\node[] at ([xshift=-0.25*\xshift em, yshift=-0.625*\yshift em]x) {\(N\)};
	
	\draw[-Latex] (x) -- (y);
	\draw[-Latex] (y) -- (yhat);
	\draw[-Latex] (x.north) to [out=30, in=150](yhat.north);
	\draw[-Latex] (y) -- (yhat);

	\draw[-Latex] (theta_y) -- (y);
	\draw[-Latex] (theta_yhat) -- (yhat);
	\draw[-Latex] (noise) -- (yhat);

	\draw[-Latex, dashed] (x.south) to [out=-30, in=-150](y.225);
	\draw[-Latex, dashed] (yhat.south) to [out=-150, in=-30](y.315);
	\draw[-Latex, dashed] (rho) -- (y.248);
\end{tikzpicture}
        \caption{Proposed Graphical Model}
        \label{fig:graphical_model}
    \end{subfigure}
    \caption{  
    (\subref{fig:motivation_graph}) Comparison of test accuracy \% (as a function of training epoch) between the original DivideMix~\cite{li2020dividemix} (solid, blue curve) and our modified DivideMix (dashed, red curve) that selects the clean and noisy data based on a fixed noise rate \( R(t) = 1-\epsilon= 50\%\) using the small-loss criterion on CIFAR100~\cite{krizhevsky2009learning} at \(0.5\) IDN~\cite{xia2020part};
    (\subref{fig:graphical_model}) The proposed probabilistic graphical model that generates noisy-label \(\hat{Y}\) conditioned on the image \(X\), the latent clean-label \(Y\) and noise rate \(\epsilon\), where forward pass (solid lines) is parameterised by \(\theta_{y}, \theta_{\hat{y}}\) and \(\epsilon\) representing the generation step, and the backward pass (dashed lines) is parameterised by \(\rho\).}
    \label{fig:fig1}
\end{figure*}

The sample selection curriculum defines a threshold to be used with one of the criteria listed above to classify the training samples as clean or noisy at each training epoch~\cite{xiao2015learning}. For example, the threshold can be fixed to an arbitrary clustering score that separates clean and noisy samples~\cite{li2020dividemix}, but this strategy does not account for the proportion of label noise in the training set, nor does it consider the dynamics of the selection of noisy-label samples during the training. 
The consideration of such dynamics has been studied in~\cite{han2018co, yao2020searching}, which defined a curriculum of the noisy-label sampling rate $R(t)$ as a function of the training epoch $t \in \{1,\dots,T\}$. The curriculum \(R(t)\) defines a sampling rate close to \(100\%\) of the training set at the beginning of the training, which is then reduced to arbitrarily low rates at the end of the training. 
In practice, the function \(R(t)\) is predefined~\cite{han2018co} or learned by weighting a set of basis functions~\cite{yao2020searching}.

Although generally effective, these techniques do not consider the label noise rate estimated from the training set, making them vulnerable to over-fitting (if too many noisy-label samples are classified as clean) or under-fitting (if informative clean-label samples are classified as noisy).
It can be argued that the estimation of the label transition matrix~\cite{yao2020dual, cheng2022instance, yang2022estimating, Chen_2023} 
aims to recover the noise rate affecting pairwise label transitions. However, label transition matrix techniques follow a quite different strategy compared to sample selection methods, where their main challenge is the general underconstrained aspect of the matrix estimation, making them sensitive to large noise rates and not scalable to a high number of classes~\cite{song2022learning}. Although several methods address label noise, none estimates the label noise rate directly from the training data to guide sample selection. Addressing this gap is the primary challenge of our study.

To underscore the importance of using the noise rate for sample selection during training, we experiment with
CIFAR100~\cite{krizhevsky2009learning} at an IDN rate \(\epsilon = 50\%\)~\cite{xia2020part} (noise rate specifications and other details are explained in~\cref{sec:experiments}). We use DivideMix~\cite{li2020dividemix} and replace its sample selection (based on an arbitrary clustering score \cite{dempster1977maximum,titterington1984on,mclachlan1996finite}) by a thresholding process that classifies the \( R(t) = 1-\epsilon = 50 \% \) largest loss samples as noisy, and the remaining ones as clean in all training epochs \(t\in\{1, \dots, T\}\). This sample selection is used for the semi-supervised learning of DivideMix~\cite{li2020dividemix}. As shown in \cref{fig:motivation_graph}, the new sample selection approach based on the \say{manually provided} noise rate  (dashed red curve) improves \(6\%\) in terms of prediction accuracy compared to the original DivideMix~\cite{li2020dividemix} (solid blue curve) which relies on arbitrary thresholding. 
Similar conclusions can be achieved with other methods that apply sample selection strategies to address the noisy-label learning problem, as shown in the experiments.

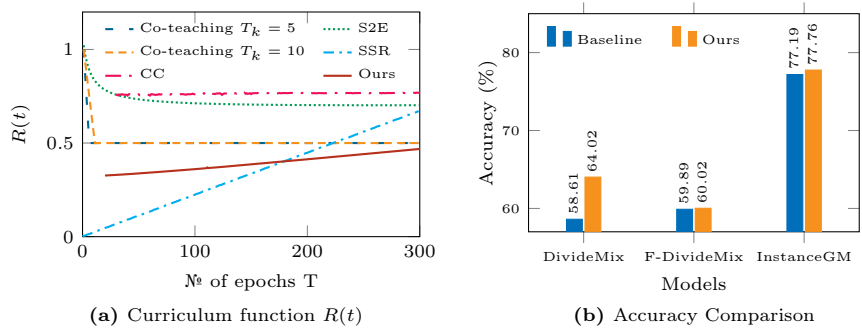
\begin{figure*}[t!]
    \centering
    \begin{subfigure}{0.49\linewidth}
        \centering
        \begin{tikzpicture}
    \pgfplotstableread[col sep=space, header=true]{images/Rt_graphs/coteaching_5.tex} \co
    \pgfplotstableread[col sep=space, header=true]{images/Rt_graphs/coteaching_10.tex} \cot
    \pgfplotstableread[col sep=space, header=true]{images/Rt_graphs/ssr.tex} \sr
    \pgfplotstableread[col sep=space, header=true]{images/Rt_graphs/s2e.tex} \se
    \pgfplotstableread[col sep=space, header=true]{images/Rt_graphs/cc.tex} \cc
    \pgfplotstableread[col sep=space, header=true]{images/Rt_graphs/ours.tex} \ours
    \begin{axis}[
        height = 0.5\linewidth,
        width = 0.75\linewidth,
        xlabel={\textnumero~of epochs T},
        xlabel style={font=\scriptsize},
        xticklabel style = {font=\scriptsize},
        xmin=0,
        xmax=300,
        ylabel={\(R(t)\)},
        ylabel style={font=\scriptsize, yshift=0em},
        yticklabel style = {font=\scriptsize},
        ymin=0,
        ymax=1.2,
        restrict x to domain=0:300,
        legend entries={Co-teaching \(T_{k} = 5\), S2E, Co-teaching \(T_{k} = 10\), SSR, CC, Ours},
        legend columns=2,
        legend style={draw=none, fill=none, yshift=0.29em, font=\tiny, /tikz/every even column/.append style={column sep=0.5em}},
        legend image post style={scale=0.625},
        legend cell align={left},
        legend pos=north east,
        scale only axis
    ]

    \addplot[mark=none, MidnightBlue, thick, loosely dashed] table[x={epochs}, y={Rt}]{\co};
    \addplot[mark=none, ForestGreen, solid, thick, densely dotted] table[x={epochs}, y={Rt}]{\se};
    \addplot[mark=none, BurntOrange, thick, densely dashed] table[x={epochs}, y={Rt}]{\cot};
    \addplot[mark=none, Cyan, thick, dash pattern=on 4pt off 2pt on 1pt off 2pt] table[x={epochs}, y={Rt}]{\sr};
    \addplot[mark=none, OrangeRed, thick, dash pattern=on 6pt off 3pt on 1pt off 3pt] table[x={epochs}, y={Rt}]{\cc};
    \addplot[mark=none, BrickRed, solid, thick, solid] table[x={epochs}, y={Rt}]{\ours};
    \end{axis}
\end{tikzpicture}
        \caption{Curriculum function $R(t)$}\label{fig:R_t}
    \end{subfigure}
    \hfill
    \begin{subfigure}{0.49\linewidth}
        \begin{tikzpicture}
  \begin{axis}[
    ybar=1pt, 
    bar width=6pt, 
    symbolic x coords={DivideMix, F-DivideMix, InstanceGM},
    xtick=data,
    ylabel={Accuracy (\%)},
    xlabel={Models},
    nodes near coords,
    nodes near coords align={vertical},
    nodes near coords style={
        font=\tiny, 
        rotate=90, 
        yshift=-1.5pt, 
        xshift=10pt, 
        inner sep=0pt,
    },
    width=\linewidth,
    height=0.75\linewidth, 
    ylabel style={font=\scriptsize, yshift=-0.5em},
    yticklabel style={font=\tiny},
    xlabel style={font=\scriptsize},
    xticklabel style={font=\tiny},
    legend style={
      draw=none,
      fill=none,
      font=\tiny,
    },
    legend pos=north west,
    legend columns=-1,
    legend style={/tikz/every even column/.append style={column sep=1em}},
    enlarge x limits=0.25,
    ymin=57,
    ymax=85
    ]
    \addplot[draw=NavyBlue, fill=NavyBlue] coordinates {
        (DivideMix, 58.61)
        (F-DivideMix, 59.89)
        (InstanceGM, 77.19)
    };
    \addplot[draw=BurntOrange, fill=BurntOrange] coordinates {
        (DivideMix, 64.02)
        (F-DivideMix, 60.02)
        (InstanceGM, 77.76)
    };
    \legend{Baseline, Ours}
  \end{axis}
\end{tikzpicture}
        \caption{Accuracy Comparison} \label{fig:bar-chart_various}
    \end{subfigure}
    \caption{  
    (\subref{fig:R_t}) Visual comparison of different \(R(t)\) on CIFAR100~\cite{krizhevsky2009learning} at \(0.5\) IDN~\cite{xia2020part}: \emph{(i)} Co-teaching~\cite{han2018co} with curriculum based on  hyper-parameter \(T_{k} \in \{5,10\}\), where \(R(t) = 1 - \tau \cdot \min(\nicefrac{t}{T_k}, 1)\) 
    with \(\tau = \epsilon = 0.5\) being manually set; \emph{(ii)}  S2E~\cite{yao2020searching}, where \(R(t)\) is estimated with a bi-level optimisation; \emph{(iii)} SSR ~\cite{feng2021ssr}, where R(t) is based on a relabelling pattern, \emph{(iv)} CC~\cite{zhao2022centrality} following the small-loss hypothesis~\cite{li2020dividemix} and cosine similarity for R(t);
    and \emph{(v)} ours.
     (\subref{fig:bar-chart_various}) Accuracy of noisy-label learning methods on CIFAR100~\cite{krizhevsky2009learning} at \(0.5\) IDN~\cite{xia2020part}, including DivideMix~\cite{li2020dividemix}, FINE~\cite{kim2021fine} and InstanceGM~\cite{garg2023instance}, without (left, blue) and (right, orange) integration of our proposed graphical model for estimating the noise rate \(\epsilon\) and sample selection rate based on \(\epsilon\). }
\end{figure*}

In this paper, we introduce a new sample selection strategy centered on estimating the label noise rate of the training set.
Our strategy is based on our novel noisy-label learning graphical model illustrated in~\cref{fig:graphical_model}.
This model can be seamlessly integrated with state-of-the-art (SOTA) noisy-label learning techniques, providing them with a precise noise rate estimate and subsequently refining the sample selection curriculum. 
In particular, our model's curriculum is not restrained by predefined \(R (t) \)  functions~\cite{han2018co,yao2020searching}, but rather relies on a dynamically estimated noise rate sourced directly from the training set, as depicted in~\cref{fig:R_t}. Our method dynamically estimates training set noise rates to mitigate overfitting and underfitting, seamlessly integrates with existing algorithms, and serves as a robust foundation for future noisy-label learning research. To summarise, our main contributions include:
\begin{itemize}
\item An innovative noisy-label learning graphical model, shown in~\cref{fig:graphical_model}, that not only estimates but also leverages the noise rate from the training dataset to produce a refined sample selection curriculum for SOTA LNL methods.
\item A simple and synergistic integration strategy of our novel graphical model with several SOTA noisy-label learning algorithms, such as DivideMix~\cite{li2020dividemix} and SSR~\cite{feng2021ssr}, to improve their sample selection effectiveness, leading to an increased test accuracy, as shown in~\cref{fig:bar-chart_various}.
\end{itemize}
Empirical evaluations show the role of our proposed sample selection methodology in improving the efficacy of leading noisy-label learning algorithms across various synthetic and real-world IDN benchmarks. Although our primary focus lies in addressing IDN problems, because of its more challenging and realistic nature, we have also applied our model to IIN problems, with detailed results in~\cref{tab:cifar_iin_supp} and discussion in~\cref{sec:further_exp_discussion_supp} of the supplementary material.
With our innovative approach, we aim to improve the accuracy of SOTA LNL methods. 

\section{Related Work}\label{sec:related_work}
It is well-known that DNNs suffer from overfitting when trained with noisy-labels~\cite{zhang2017understanding}, resulting in poor generalisation~\cite{liu2020early, zhang2021learning_}. 
To mitigate the problems related to noisy-label learning, several techniques have been developed, including \textit{noise-robust loss functions}~\cite{ma2020normalized}, \textit{noise-label sample selection}~\cite{li2020dividemix, zheltonozhskii2022contrast,Tanaka_2018_CVPR, kim2021fine, zhao2022centrality}, \textit{meta-learning}~\cite{zhang2021learning_meta}, and \textit{re-labeling} followed by \textit{sample selection}~\cite{feng2021ssr}.  These techniques can handle many types of label noise (e.g., symmetric and asymmetric), but the most challenging type, i.e., IDN~\cite{song2022learning}, tends to be successfully addressed with methods that have a
training stage based on sample selection techniques~\cite{li2020dividemix, zheltonozhskii2022contrast,Tanaka_2018_CVPR, kim2021fine, zhao2022centrality}. These sample selection techniques separate clean and noisy-label samples, where clean samples are treated as labelled, and noisy samples are discarded~\cite{han2018co, jiang2018mentornet, yao2020searching,cheng2020learning_iclr} or treated as unlabeled samples~\cite{li2020dividemix, zheltonozhskii2022contrast, nishi2021augmentation, garg2023instance, cordeiro2021propmix} for semi-supervised learning~\cite{berthelot2019mixmatch}. Another solution consists of aligning clean and noisy samples with an information fusion method~\cite{jiang2022an}.
A major limitation of these approaches is the overly simplistic design of the curriculum to select clean and noisy-label samples during training. 
It is either predetermined~\cite{jiang2018mentornet, han2018co} or learned from a set of predetermined basis functions~\cite{yao2020searching}. 
We argue that the design of a curriculum based on an estimated noise rate would benefit such methods, as motivated by Fig.~\ref{fig:motivation_graph}, but to the best of our knowledge, such estimation has not been explored by previous methods.
The closest idea explored is based on the estimation of the type of noise instead of the noise rate~\cite{xiao2015learning}. It is possible to argue that the estimation of the label transition matrices~\cite{xia2020part, cheng2022instance, Zhu_2021_CVPR, chen2021beyond}, composed of the learned noise rate that affects the transition between pairs of labels, is a way to estimate the label noise rate. 
However, they portray results of comparatively lower accuracy for large real-world datasets or for IDN problems~\cite{xia2020part,song2022learning}. 
A possible reason for these poorer results is that label transition approaches suffer from identifiability issues~\cite{frenay2013classification}, where any clean-label distribution assignment is acceptable as long as the distribution of observed labels can be reconstructed~\cite{frenay2014comprehensive}. 
This makes the identification of the actual underlying clean-labels challenging. One solution is to consider the use of multiple annotations per sample to help analyse agreements and disagreements for improved identification of clean patterns~\cite{malach2017decoupling, frenay2013classification}. However,
the annotation of datasets with a single label per training sample is already challenging; the complexity is significantly larger to acquire  multiple labels per sample.

An alternative technique to handle IDN problems is based on graphical models that represent the relationship between various observed and latent variables~\cite{garg2023instance, yao2021instance, xiao2015learning}. 
The approaches in~\cite{garg2023instance, yao2021instance} use graphical models that rely on a generative approach to produce noisy-labels from the respective image features and latent clean-labels. However, previous graphical models do not take into account the underlying noise rate parameter during modeling.
Our work is the first graphical model approach to estimate the noise rate of the dataset.  
In addition, a close examination of existing literature reveals a trend: when traditional graphical models are integrated with SOTA noisy-label learning methods, the outcomes often fall short of expectations in terms of accuracy~\cite{yao2021instance, garg2023instance, bae2022noisy}. This observation underscores a strong point of our work. Our approach is uniquely tailored for easy integration with current SOTA noisy-label learning methods for sample selection, which, in turn, improves over the SOTA classification accuracy.

\section{Method}\label{sec:method}
In this section, we present our new graphical model that estimates the noise rate, which will be used in the sample selection process.
Let \(\mathcal{D} = \{(x_i, \hat{y}_i)\}_{i=1}^{N}\) be the noisy-label training set containing \(d\)-dimensional data vector \(x_i \in \mathcal{X} \subseteq \mathbb{R}^d\) 
and its respective \(C\)-dimensional one-hot encoded observed (potentially corrupted) label \( \hat{y}_{i} \in \hat{\mathcal{Y}} = \{\hat{y}: \hat{y} \in \{0, 1\}^{C} \wedge \mathbf{1}_{C}^{\top} \hat{y} = 1 \} \), where \( \mathbf{1}_{C} \) is a vector of ones with $C$ dimensions.
The aim is to estimate the label noise rate \(\epsilon\), used for the generation of noisy-label training data from the observed training dataset \(\mathcal{D}\) and integrate this label noise rate into the sample selection strategy. 

\subsection{Graphical Model}\label{subsec:noiseRateestimation_sub}
We portray the generation of noisy-label via the probabilistic graphical model shown in~\cref{fig:graphical_model}.
The observed random variables, denoted by shaded circles, are data \(X\) and the corresponding noisy-label \(\hat{Y}\). We also have one latent variable, namely: the clean-label \(Y\).  
Under our proposed modeling assumption, a noisy-label of a data instance can be generated as follows:

\begin{itemize}
    \item sample an instance from \(p(X)\), i.e. : \(x \sim p(X)\)
    \item sample a clean-label from the clean-label distribution: \(y \sim \mathrm{Categorical}(Y; f_{\theta_{y}}(x))\)
    \item sample a noisy-label from the noisy-label distribution: \\\(\hat{y} \sim \mathrm{Categorical}(\hat{Y}; \epsilon \times f_{\theta_{\hat{y}}}(x,f_{\theta_{y}}(x)) + (1 - \epsilon) \times y)\),
\end{itemize}
where \(\mathrm{Categorical}(.)\) denotes a categorical distribution, \(f_{\theta_{y}}: \mathcal{X} \to \Delta_{C - 1}\) and \(f_{\theta_{\hat{y}}}: \mathcal{X} \times \Delta_{C - 1} \to \Delta_{C - 1}\) denote two classifiers for the clean-label \(Y\) and noisy-label \(\hat{Y}\), 
respectively, with \(\Delta_{C - 1} = \{s: s \in [0, 1]^{C} \wedge \pmb{1}_{C}^{\top} s = 1\}\) being the \((C - 1)\)-dimensional probability simplex. According to the data generation process shown in~\cref{fig:graphical_model}, \(\epsilon\) corresponds to \(\mathbb{E}_{(x, \hat{y}) \sim p(X, \hat{Y})} [P(\hat{y}\neq y|x) ]\), which is the label noise rate of the training dataset of interest.
Our aim is to infer the parameters \(\theta_{y}, \theta_{\hat{y}}\) and \(\epsilon\) from a noisy-label dataset \(\mathcal{D}\) by maximising the following log-likelihood:
\begin{equation}\label{eq:objective}
\scalebox{0.95}{$
\begin{aligned}[b]
    & \max_{\theta_{y}, \theta_{\hat{y}}, \epsilon} \mathbb{E}_{(x_{i}, \hat{y}_{i}) \sim \mathcal{D}} \left[ \ln p(\hat{y}_{i} | x_{i}; \theta_{y}, \theta_{\hat{y}}, \epsilon) \right] = \max_{\theta_{y}, \theta_{\hat{y}}, \epsilon} \mathbb{E}_{(x_{i}, \hat{y}_{i}) \sim \mathcal{D}} \left[ \ln \textstyle \sum_{y_{i}} p(\hat{y}_{i}, y_{i} | x_{i}; \theta_{y}, \theta_{\hat{y}}, \epsilon) \right].
\end{aligned}
$}
\end{equation}
Due to the presence of the clean-label \(y_{i}\), it is difficult to directly evaluate the log-likelihood in \cref{eq:objective}. Therefore, we employ the \emph{expectation - maximisation} (EM) algorithm~\cite{dempster1977maximum} to maximise the log-likelihood. The main idea of the EM algorithm is to \emph{(i)} construct a tight lower bound of the likelihood in \cref{eq:objective} by estimating the posterior of the latent variable \(Y\) (known as \emph{expectation step}) and \emph{(ii)} maximise that lower bound (known as \emph{maximisation step}). Formally, let \(q(y_{i} | x, \hat{y}; \rho)\) be an arbitrary distribution over a clean-label \(y_{i}\). The evidence lower bound (ELBO) of the log-likelihood in \cref{eq:objective} can be obtained through Jensen's inequality and is presented as follows: 
\begin{equation}\label{eq:lower_bound}
\begin{aligned}[b]
Q(\theta_{y}, \theta_{\hat{y}}, \epsilon, \rho) & = \mathbb{E}_{(x_{i}, \hat{y}_{i}) \sim \mathcal{D}} \left[ \mathbb{E}_{q(y_{i} | x_{i}; \hat{y}_{i}; \rho)} \left[ \ln p(\hat{y}_{i} | x_{i}, y_{i};  \theta_{y}, \theta_{\hat{y}}, \epsilon) \right] \right. \\
& \qquad \left. - \mathrm{KL} \left[ q(y_{i} | x_{i}, \hat{y}_{i}; \rho) \Vert p(y_{i} | x_{i}; \theta_{y}, \theta_{\hat{y}}, \epsilon) \right] \right] \\
& = \mathbb{E}_{(x_{i}, \hat{y}_{i}) \sim \mathcal{D}} \left[ \mathbb{E}_{q(y_{i} | x_{i}; \hat{y}_{i}; \rho)} \left[ \ln p(\hat{y}_{i} | x_{i}, y_{i}; \theta_{\hat{y}}, \epsilon) \right. \right. \\
& \qquad \left.\left. + \ln p(y_{i} | x_{i}; \theta_{y}) \right] + \mathbb{H} \left[ q(y_{i} | x_{i}, \hat{y}_{i}; \rho) \right] \right],
\end{aligned}
\end{equation}
where \(\mathrm{KL}[q \Vert p]\) is the Kullback – Leibler divergence between distributions \(q\) and \(p\) and \(\mathbb{H}(q)\) is the entropy of the distribution \(q\). The EM algorithm is then carried out iteratively by alternating the following two steps:

\paragraph{E step:} we maximise the ELBO in \eqref{eq:lower_bound} w.r.t. \(q(y_{i} | x_{i}, \hat{y}_{i}; \rho)\). Theoretically, such an optimisation results in \(\mathrm{KL} \left[ q(y_{i} | x_{i}, \hat{y}_{i}; \rho) \Vert p(y_{i} | x_{i}, \hat{y}_{i}) \right] = 0\) or \(q(y_{i} | x_{i}, \hat{y}_{i}; \rho) = p(y_{i} | x_{i}, \hat{y}_{i})\). This is equivalent to estimating the posterior of the clean-label \(y_{i}\) given noisy-label data \((x_{i}, \hat{y}_{i})\). Obtaining the posterior \(p(y_{i} | x_{i}, \hat{y}_{i})\) is, however, intractable for most deep-learning models. To mitigate such an issue, we follow the \emph{variational EM} approach~\cite{neal1998view} by employing an approximate posterior \(q(y_{i} | x_{i}, \hat{y}_{i}; \rho^{(t)})\) that is the closest to the true posterior \(p(y_{i} | x_{i}, \hat{y}_{i})\), where:
\begin{equation}\label{eq:e_step}
    \rho^{(t)} = \arg\max_{\rho} Q \left( \theta_{y}^{(t)}, \theta_{\hat{y}}^{(t)}, \epsilon^{(t)}, \rho \right),
\end{equation}
with the superscript \({}^{(t)}\) denoting the parameters at the \(t\)-th iteration. Although this results in a non-tight lower bound of the log-likelihood in \cref{eq:objective}, it does increase the variational bound \(Q\).

\paragraph{M step:} we maximise the ELBO in \cref{eq:lower_bound} w.r.t. \(\theta_{y}, \theta_{\hat{y}}\) and \(\epsilon\) given \(\rho^{(t)}\) obtained in the E step:
\begin{equation}
    \theta_{y}^{(t + 1)}, \theta_{\hat{y}}^{(t + 1)}, \epsilon^{(t + 1)} = \arg\max_{\theta_{y}, \theta_{\hat{y}}, \epsilon} Q \left( \theta_{y}, \theta_{\hat{y}}, \epsilon, \rho^{(t)} \right).
    \label{eq:m_step}
\end{equation}
The estimated noise rate \(\epsilon\) can then be integrated into certain noisy-label algorithms to train the models of interest as mentioned in \cref{sec:introduction}. In addition, the inference of noise rate \(\epsilon\) might be associated with the identifiability issue when estimating the clean-label \(Y\)~\cite{liu2022identifiability}, i.e., there exists multiple sets of \(\rho\) and \(\theta_{y}\), where each set can explain the observed noisy-label data equally well. Such issues are addressed in the following subsection.

\begin{algorithm}[t]
    \footnotesize
    \caption{Proposed noisy-label learning  that relies on the estimation of noise rate \(\epsilon\) to build a sample selection curriculum.}
    \label{algorithm:Proposed_Algo}
    \begin{minipage}{\linewidth}
    \begin{algorithmic}[1]
        \Procedure{Noise rate estimation and integration}{$\mathcal{D}, T, \lambda$}
            \LComment{$\mathcal{D} = \{(x_i, \hat{y}_i)\}_{i=1}^{N}$: training set with noisy-label data}
            \LComment{$T$: number of epochs}
            \LComment{$\lambda$: a hyper-parameter}
            \LComment{Notation: $\Theta^{(t)} = (\theta_{y}^{(t)}, \theta_{\hat{y}}^{(t)}, \epsilon^{(t)}) $}
            \State Initialise $\theta^{(1)}_y, \theta^{(1)}_{\hat{y}}, \epsilon^{(1)}$ and $\rho^{(0)}$
            \State $\theta_y^{(1)} \gets$ \Call{Warm up}{$\mathcal{D}, \theta_{y}^{(1)} $}
            \State $t \gets 0$
            \For{$n_{\mathrm{epoch}} = 1:T$}
                \For{each mini-batch $\mathcal{S}$ in \Call{shuffle}{$\mathcal{D}$}}
                    \State $t \gets t + 1$
                    \Statex
                    \State $\mathcal{S}_{\mathrm{clean}}, \mathcal{S}_{\mathrm{noisy}} \gets$ \Call{Sample Selection}{$ \mathcal{S}, \theta_{y}^{(t)}, \epsilon^{(t)} $} \Comment{\cref{eq:sample_selection}}
                    \Statex
                    \LComment{Variational E-step as in \cref{eq:e_step}}
                    \State $\rho^{(t)} \gets \Call{E step}{\mathcal{S}, \theta_{y}^{(t)}, \theta_{\hat{y}}^{(t)}, \epsilon^{(t)}, \rho^{(t - 1)}}$
                    \Statex
                    \LComment{M-step as in \cref{eq:m_step_2}}
                    \State $\theta_{y}^{(t + 1)}, \theta_{\hat{y}}^{(t + 1)}, \epsilon^{(t + 1)} \gets \Call{M step}{\mathcal{S}_{\mathrm{clean}}, \mathcal{S}_{\mathrm{noisy}}, \theta_{y}^{(t)}, \theta_{\hat{y}}^{(t)}, \epsilon^{(t)}, \rho^{(t)}, \lambda}$
                \EndFor
                \Statex
            \EndFor
            \State \Return $\theta_{y}$ \Comment{parameter of the clean-label classifier}
        \EndProcedure
    \end{algorithmic}
    \end{minipage}
\end{algorithm}

\subsection{Sample Selection} \label{sec:sample_selection}

The identifiability issue when inferring the clean-label \(Y\) from noisy-label data \((X, \hat{Y})\) can be mitigated either by acquiring multiple noisy-labels~\cite{liu2022identifiability} or introducing additional constraints, such as small loss hypothesis~\cite{han2018co} or FINE~\cite{kim2021fine}. 
Since requesting additional noisy-labels per training sample is not always possible, we follow the latter approach by imposing a constraint, denoted as \(L(\theta_{y}, \epsilon^{(t)})\), over \(\theta_{y}\) in the M step via a sample selection approach based on the estimated noise rate \(\epsilon^{(t)}\). Formally, we propose a new curriculum when selecting samples as follows:
\begin{equation}
    R(t) = 1-\epsilon^{(t)}.
    \label{eq:curriculum}
\end{equation}
In the simplest case, such as Co-teaching~\cite{han2018co} or FINE~\cite{kim2021fine}, the constraint for \(\theta_{y}\) can be written as:
\begin{equation}\label{eq:loss_SOTA}
    \begin{aligned}[b]
    & L(\theta_{y}, \epsilon^{(t)}) = \sum_{(x_{i}, \hat{y}_{i}) \in \mathcal{S}_{\text{clean}}} \mathrm{KL} \left[ \mathrm{Categorical}(Y; \hat{y}) \Vert \mathrm{Categorical}(Y; f_{\theta_{y}}(x_{i})) \right],  
    \end{aligned}
\end{equation}
where:
\begin{equation}\label{eq:sample_selection}
    \begin{aligned}[b]
        & \mathcal{S}_{\mathrm{clean}} = \left\{ (x_{i}, \hat{y}_{i}): (x_{i}, \hat{y}_{i}) \in \mathcal{D} \wedge z(x_{i}, y_{i}) \le m \} \right\},\\
        & m \in \{ m: \Pr(z(x_{i}, y_{i}) \le m) \ge R(t) \wedge \Pr(z(x_{i}, y_{i}) \ge m) \ge 1 - R(t) \}, \\
        & \mathcal{S}_{\mathrm{noisy}} = \mathcal{D} \setminus \mathcal{S}_{\mathrm{clean}}.
    \end{aligned}
\end{equation}
with \(R(t)\) defined in~\eqref{eq:curriculum},
and \(z(x, y)\) representing the score of a criterion (e.g., loss~\cite{li2020dividemix,zheltonozhskii2022contrast, zhao2022centrality}, distance to the largest eigenvectors~\cite{kim2021fine}, or KNN scores~\cite{feng2021ssr}).
Intuitively, the loss in \cref{eq:loss_SOTA} is simply the cross-entropy loss on the \(\lfloor R(t) \times N \rfloor\) samples that have smallest scores (with \(\lfloor . \rfloor\) being the floor function). One can also extend to other SOTA models by replacing the loss \(L\) accordingly. For example, if DivideMix~\cite{li2020dividemix} is used as a base model to constrain \(\theta_{y}\), \(L\) will include two additional terms: loss on un-labeled data and regularisation using \emph{mixup}~\cite{zhang2017mixup}.

\subsection{Training and Testing}\label{subsec:noise_rate_integration}

\begin{figure*}[t]
    \centering
    \begin{adjustbox}{max width=0.8\textwidth}
        \input{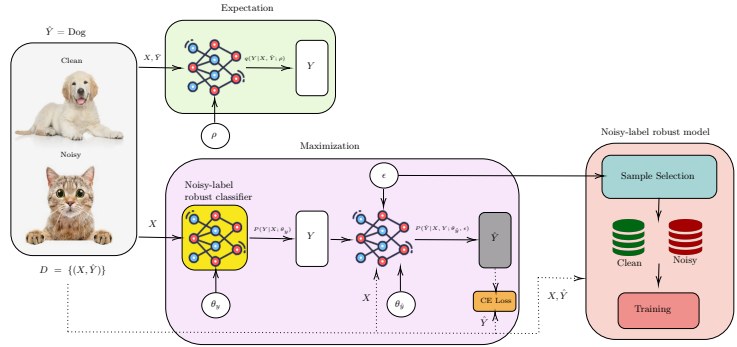}
    \end{adjustbox}
    \caption{Our training algorithm uses an existing noisy-label classifier (e.g., DivideMix~\cite{li2020dividemix}) parameterised by \(\theta_{y}\) as the clean-label model \( p(Y|X;\theta_{y}) \).
    The generation of noisy-label (given \(X\) and \(Y\)) is carried out by a model parameterised by noise rate \(\epsilon\) and \(\theta_{\hat{y}}\). 
    The noisy-label classifier is based on a sample-selection mechanism that uses a curriculum \(R(t)=1 - \epsilon^{(t)}\). 
    }
    \label{fig:methodology}
\end{figure*}

Given the sample selection approach in \cref{sec:sample_selection},
the M step in \cref{eq:m_step} is slightly modified by integrating the constraint in \cref{eq:loss_SOTA}, which can be written as 
\begin{equation}
    \begin{aligned}[b]
         \theta_{y}^{(t + 1)}, \theta_{\hat{y}}^{(t + 1)},  \epsilon^{(t + 1)} & = \arg\max_{\theta_{y}, \theta_{\hat{y}}, \epsilon} Q \left( \theta_{y}, \theta_{\hat{y}}, \epsilon, \rho^{(t)} \right) - \lambda \, L(\theta_{y}, \epsilon^{(t)}),
    \end{aligned}
    \label{eq:m_step_2}
\end{equation}
where \(\lambda\) is a hyper-parameter and \(L\) is defined in \cref{eq:loss_SOTA}.
The training procedure is summarised in \cref{algorithm:Proposed_Algo} and visualised in \cref{fig:methodology}. In the implementation, we integrate the proposed method into existing models, such as DivideMix~\cite{li2020dividemix} or FINE~\cite{kim2021fine}. Note that the clean-label classifier \(f_{\theta_{y}}(.)\) is also the clean classifier of the base model. 

\section{Experiments}\label{sec:experiments}
We show extensive experiments in several noisy-label synthetic benchmarks with CIFAR100~\cite{krizhevsky2009learning}, and real-world benchmarks, including CNWL's red mini-ImageNet~\cite{jiang2020beyond}, Clothing1M~\cite{xiao2015learning} and mini-WebVision~\cite{li2017webvision}. 
~\cref{subsec:implementation} describes implementation details. 
We evaluate our approach by plugging SOTA models into \( p(y|x;\theta_y) \), defined in~\cref{sec:method}, with results being shown in~\cref{subsec:comparison}, 
and ablation studies in~\cref{subsec:ablation}.

\subsection{Dataset Descriptions}\label{subsec:dataset_supp}
We follow the existing literature to generate the IDN labels~\cite{xia2020part} for \textbf{CIFAR100}~\cite{krizhevsky2009learning}. 
The dataset contains \(50,000\) training images and \(10,000\) testing images of shape \(32 \times 32 \times 3\). The dataset is class-balanced with 100 different categories and the 
IDN rates considered are {\(0.2, 0.3, 0.4 \text{ and } 0.5\)}~\cite{xia2020part}.
Another important IDN benchmark dataset is the \textbf{red mini-ImageNet}, a real-world dataset from CNWL~\cite{jiang2020beyond}. 
In this dataset, there are \(100\) distinct classes, with each class containing \(600\) images.
The images and their corresponding noisy-labels have been crawled from the internet at various controllable label noise rates.
For a fair comparison with the existing literature~\cite{xu2021faster, garg2023instance, cordeiro2021propmix}, we have resized the images to \(32 \times 32\) pixels from the \(84 \times 84\) original pixel settings. We show results with \(40\%, 60\% \text{ and } 60\%\)  noise rates on this dataset.
\textbf{Clothing1M}~\cite{xiao2015learning} is a real-world clothing classification dataset that contains 1 \emph{million} images with an estimated noise rate of \(38.5\%\). 
The dataset contains \(14\) different categories, and the labels are generated from surrounding texts. 
Images in this dataset are of different sizes, and we follow the resize structure suggested in~\cite{li2020dividemix, kim2021fine}. This dataset also contains \(50k\) manually validated clean training images, \(14k\) images for the validation set, and \(10k\) testing images. 
We have not used the clean training, validation set, or any extra training images while training. Only the testing set is used for evaluation.
\textbf{Mini-WebVision}~\cite{li2020dividemix} contains \(65,944\) images with their respective labels from the \(50\) different initial categories from the WebVision~\cite{li2017webvision}, with image size of \(256 \times 256\) pixels. Following the evaluation process commonly used in this benchmark, \(50\) categories from the ILSVRC12~\cite{krizhevsky2012imagenet} dataset are also used for testing.

\subsection{Implementation}\label{subsec:implementation}
All  methods are implemented in Pytorch~\cite{paszke2019pytorch} and use one NVIDIA RTX 3090 card for training and testing. 
As mentioned in the original papers, hyperparameter settings are kept the same for the baselines used in the proposed algorithm. 
All classifier architectures are also kept the same as the baseline models. 
A random initialisation of noise rate parameter \(\epsilon\) with the sigmoid as its activation function is employed for all experiments to maintain the fairness of the comparisons with other approaches.
The value of \(\lambda\) in~\cref{eq:m_step_2} is set to \(1\) for all the cases. We integrate many SOTA approaches~\cite{li2020dividemix, zheltonozhskii2022contrast, garg2023instance, kim2021fine, feng2021ssr, zhao2022centrality} into our graphical model, as explained in~\cref{subsec:noise_rate_integration}. For CIFAR100~\cite{krizhevsky2009learning} with IDN~\cite{xia2020part}, we integrate DivideMix~\cite{li2020dividemix}, C2D~\cite{zheltonozhskii2022contrast}, CC~\cite{zhao2022centrality}, and InstanceGM~\cite{garg2023instance} into our model, given their superior  performance across various noise rates. Additionally, we also use F-Dividemix from FINE~\cite{kim2021fine}, as shown in~\cref{fig:bar-chart_various}. 
For red mini-ImageNet~\cite{jiang2020beyond}, we test our proposed approach with and without DINO self-supervision~\cite{caron2021emerging}. For the implementation without self-supervision, we use DivideMix~\cite{li2020dividemix} and InstanceGM~\cite{garg2023instance}, and for the self-supervised version, we only use InstanceGM~\cite{garg2023instance}. 
The models for Clothing1M~\cite{xiao2015learning} and mini-WebVision~\cite{li2020dividemix} (validation on ImageNet~\cite{krizhevsky2012imagenet}) are trained using DivideMix~\cite{li2020dividemix}, SSR~\cite{feng2021ssr}, C2D~\cite{zheltonozhskii2022contrast} and CC~\cite{zhao2022centrality}. Additional implementation details, empirical analysis, and computational analysis are presented in the supplementary material in \cref{sec:implementation_supp,sec:empirical_analysis,sec:computational_analysis}, respectively.

\subsection{Comparison with SOTA}\label{subsec:comparison}
This section compares our approach with SOTA methods on datasets with the IDN settings and noisy real-world settings. The bold text in tables indicate SOTA results, and our results are in the greyed rows.

\subsubsection{Synthetic Instance-Dependent Noise}\label{subsubsec:idn}

The comparison between various baselines and our proposed method on CIFAR100~\cite{krizhevsky2009learning} with IDN~\cite{xia2020part} is shown in~\cref{table:cifar100} with the noise rate ranging from \(0.2\) to \(0.5\). 
It is worth noting that using our proposed model with DivideMix~\cite{li2020dividemix}, C2D~\cite{zheltonozhskii2022contrast}, CC~\cite{zhao2022centrality} and InstanceGM~\cite{garg2023instance} improve their performance in almost all  cases, particularly with large noise rates.
~\cref{table:cifar100} also shows the final noise rate \(\epsilon\) estimated by our model which is reasonably close to the simulated noise rate displayed in the table's header. 
\begin{table*}[t]
    \centering
    \caption{\emph{(left)} Test accuracy \% and \emph{(right)} final estimated noise rate \(\epsilon\) on CIFAR100~\cite{krizhevsky2009learning} under different IDN~\cite{xiao2015learning}. Other models' results are from~\cite{garg2023instance,cheng2022instance}. Here, we integrate DivideMix~\cite{li2020dividemix}, C2D~\cite{zheltonozhskii2022contrast}, CC~\cite{zhao2022centrality} and InstanceGM~\cite{garg2023instance} into our proposed model.}
    \label{table:cifar100}
    \begin{subtable}[t]{0.5\linewidth}
        \centering
        \scalebox{0.8}{
        \begin{tabular}{l c c c c}
            \toprule
            \multirow{2}{*}{\bfseries Method } & \multicolumn{4}{c}{\bfseries Noise Rates - IDN} \\
            \cmidrule{2 - 5} 
            & \textbf{0.2} & \textbf{0.3} & \textbf{0.4} & \textbf{0.5} \\
            \midrule
            CE~\cite{yao2021instance} & 30.42 & 24.15 & 21.45 & 14.42 \\
            PartT~\cite{xia2020part} & 65.33 & 64.56 & 59.73 & 56.80\\
            kMEIDTM~\cite{cheng2022instance} & 69.16 & 66.76 & 63.46 & 59.18\\
            \midrule
            DivideMix~\cite{li2020dividemix} & 77.03 & 76.33 & 70.80 & 58.61\\
            \rowcolor{Gray!25}\textbf{DivideMix-Ours} & 77.42 & 77.21 & 72.41 & 64.02\\
            \midrule
            C2D~\cite{zheltonozhskii2022contrast} & 78.61 & 78.18 & 72.89 & 63.19\\
            \rowcolor{Gray!25}\textbf{C2D-Ours} & 79.07 & 78.59 & 73.31 & 65.28  \\
            \midrule
            CC~\cite{zhao2022centrality} & 79.61 & 77.56 & 76.58 & 63.19\\
            \rowcolor{Gray!25}\textbf{CC-Ours} & \textbf{79.72} & 78.71 & 77.38 & 67.53\\
            \midrule
            InstanceGM~\cite{garg2023instance} & 79.69 & 79.21 & 78.47 & 77.19\\
            \rowcolor{Gray!25} \textbf{InstanceGM-Ours} & 79.61 & \textbf{79.40} & \textbf{79.52} & \textbf{78.21}\\
            \bottomrule
        \end{tabular}}
    \end{subtable}
    \hfill
\begin{subtable}[t]{.45\linewidth}
        \centering
        \scalebox{0.8}{
        \begin{tabular}{l c c c c}
            \toprule
            \multicolumn{4}{l}{\bfseries Estimated noise rates} \\
            \midrule
            \multirow{2}{*}{\bfseries Method} & \multicolumn{4}{c}{\bfseries Actual noise rate} \\
            \cmidrule{2-5}
            & \bfseries 0.2 & \bfseries 0.3 & \bfseries 0.4 & \bfseries 0.5 \\
            \midrule
            DivideMix-Ours & 0.18 & 0.34 & 0.47 & 0.53 \\
            C2D-Ours & 0.19 & 0.33 & 0.38 & 0.52 \\
            CC-Ours &  0.21 & 0.34 & 0.41 & 0.54\\
            InstanceGM-Ours & 0.23 & 0.37 & 0.42 & 0.47 \\
            \bottomrule
        \end{tabular}}
    \end{subtable}

\end{table*}

\begin{table*}[t]
    \centering
    \caption{\emph{(left)} Test accuracy \% and \emph{(right)} final estimated noise rate \(\epsilon\) for red mini-ImageNet~\cite{jiang2020beyond}. Other methods' results are reported in~\cite{garg2023instance,xu2021faster}. We present the results with and without self-supervision~\cite{caron2021emerging}. We integrate DivideMix~\cite{li2020dividemix} and InstanceGM~\cite{garg2023instance} into our model, with the latter tested with and without self-supervision.}
    \label{table:redmini}
    \begin{subtable}[t]{.5\linewidth}
        \centering
        \scalebox{0.8}{
        \begin{tabular}{l c c c}
            \toprule
            \multirow{2}{*}{\bfseries Method} & \multicolumn{3}{c}{\bfseries Noise rate} \\
            \cmidrule{2-4}
             & \bfseries 0.4 & \bfseries 0.6 & \bfseries 0.8 \\
            \midrule
            CE~\cite{xu2021faster} &  42.70 & 37.30 & 29.76 \\
            MixUp~\cite{zhang2017mixup} &  46.40 & 40.58 & 33.58 \\
            MentorMix~\cite{jiang2020beyond} &  47.14 & 43.80 & 33.46 \\
            FaMUS~\cite{xu2021faster} & 51.42 &  45.10 & 35.50\\
            \midrule
            DivideMix~\cite{li2020dividemix} & 46.72 & 43.14 & 34.50 \\
            \rowcolor{Gray!25}\textbf{DivideMix-Ours} & 50.70 & 45.11 & 37.44 \\ 
            \midrule
            InstanceGM~\cite{garg2023instance} &  52.24 & 47.96 & 39.62 \\
            \rowcolor{Gray!25}\textbf{InstanceGM-Ours} &  \textbf{56.61} & \textbf{51.40} & \textbf{43.83} \\
            \midrule
            \midrule
            \multicolumn{4}{l}{\bfseries With self-supervised learning}\\
            \midrule
            PropMix~\cite{cordeiro2021propmix} & 56.22 & 52.84 & 43.42\\
            \midrule
            InstanceGM-SS~\cite{garg2023instance} & 56.37 & 53.21 & 44.03\\
            \rowcolor{Gray!25}\textbf{InstanceGM-SS-Ours} & \textbf{58.29} & \textbf{53.60} & \textbf{45.47}\\
            \bottomrule
        \end{tabular}}
    \end{subtable}
    \hfill
    \begin{subtable}[t]{.45\linewidth}
        \centering
        \scalebox{0.8}{
        \begin{tabular}{l c c c}
            \toprule
            \multicolumn{4}{l}{\bfseries Estimated noise rates} \\
            \midrule
            \multirow{2}{*}{\bfseries Method} & \multicolumn{3}{c}{\bfseries Actual noise rate} \\
            \cmidrule{2-4}
            & \bfseries 0.4 & \bfseries 0.6 & \bfseries 0.8 \\
            \midrule
            DivideMix-Ours & 0.39 & 0.58 & 0.73 \\
            InstanceGM-Ours & 0.38 & 0.55 & 0.74 \\
            InstanceGM-SS-Ours & 0.48 & 0.53 & 0.69 \\
            \bottomrule
        \end{tabular}}
    \end{subtable}
\end{table*}
\begin{table*}[t]
        \centering
        \caption{Test accuracy (\%) of competing methods, and final estimated noise rate $\epsilon$ on Clothing1M~\cite{xiao2015learning}.
        Also, we have not considered competing models that rely on a clean or validation set whilst training.
        We integrate DivideMix~\cite{li2020dividemix}, C2D~\cite{zheltonozhskii2022contrast}, SSR~\cite{feng2021ssr} and CC~\cite{zhao2022centrality} into our model. DivideMix~\cite{li2020dividemix} and CC~\cite{zhao2022centrality} shows the locally reproduced results. Mentioned noise rate by~\cite{xiao2015learning} is \(0.385\). 
        }
        \label{table:clothing1M}
        \scalebox{0.8}{
        \begin{tabular}{l c c}
            \toprule
            \bfseries Method & \bfseries Test accuracy (\%) & \bfseries Estimated noise rate \\
            \midrule
            OT-Filter~\cite{feng2023ot} & 74.50 \\
            ELR+ with C2D~\cite{zheltonozhskii2022contrast} & 74.58\\
            AugDesc~\cite{nishi2021augmentation} & 75.11\\
            NCE~\cite{li2022neighborhood} & 75.30\\
            \hline
            DivideMix~\cite{li2020dividemix} & 74.32\\
            \rowcolor{Gray!25}\textbf{DivideMix-Ours} & 74.41 & 0.41\\
            \hline
            C2D~\cite{zheltonozhskii2022contrast} & 74.58 \\
            \rowcolor{Gray!25}\textbf{C2D-Ours} & 74.71 & 0.41 \\
            \hline
            SSR (class-imbalance)~\cite{feng2021ssr} & 74.12 \\
            \rowcolor{Gray!25}\textbf{SSR-Ours}& 74.20 & 0.42\\
            \hline
            CC~\cite{zhao2022centrality} & 75.24 \\
            \rowcolor{Gray!25}\textbf{CC-Ours}& \textbf{75.31} & 0.41\\
            \bottomrule
        \end{tabular}}
\end{table*}
\begin{table*}[t]
    \caption{Test accuracy (\%) and final estimated noise rate \(\epsilon\) on mini-WebVision~\cite{li2020dividemix} and validation on ImageNet~\cite{krizhevsky2012imagenet}. We integrate DivideMix~\cite{li2020dividemix}, Contrast-to-Divide (C2D)~\cite{zheltonozhskii2022contrast}, SSR~\cite{feng2021ssr} and  CC~\cite{zhao2022centrality} into our model, whilst models suffixed with \textbf{-Ours} denotes our proposed approach. ImageNet~\cite{krizhevsky2012imagenet} is only considered for validation.
    }
    \label{table:miniWebvision}
    \centering
    \scalebox{0.75}{
    \begin{tabular}{l c c c c c}
        \toprule
        \multirow{2}{*}{\bfseries Dataset} & \multicolumn{2}{c}{\bfseries mini-WebVision} & \multicolumn{2}{c}{\bfseries ImageNet} & \multirow{2}{*}{\bfseries Estimated noise rate} \\
        \cmidrule(lr){2-3} \cmidrule(lr){4-5} 
        & \textbf{Top-1} & \textbf{Top-5} & \textbf{Top-1} & \textbf{Top-5} & \\
        \midrule
        BtR~\cite{smart2023bootstrapping} & 80.88 & 92.76 & 75.96 & 92.20 & \\
        NCE~\cite{li2022neighborhood} & 79.50 & \textbf{93.80} & 76.30 & 94.10 & \\
        \midrule
        DivideMix~\cite{li2020dividemix}  & 77.32 & 91.64 & 75.20 & 91.64 & \\
        \rowcolor{Gray!25}\textbf{DivideMix-Ours}  & 78.51 & 92.03 & 76.11 & 93.24 & 0.43\\
        \midrule
        C2D~\cite{zheltonozhskii2022contrast} & 79.42 & 92.32 & 78.57 & 93.04 & \\
        \rowcolor{Gray!25}\textbf{C2D-Ours}  & 80.20 & 92.82 & \textbf{79.16} & 93.12 & 0.43 \\
        \midrule
        SSR~\cite{feng2021ssr} & 80.92 & 92.80 & 75.76 & 91.76 & \\
        \rowcolor{Gray!25}\textbf{SSR-Ours} & \textbf{81.68} & \textbf{93.80} & 76.91 & 93.05 & 0.43\\
        \midrule
        CC~\cite{zhao2022centrality} & 79.36 & 93.64 & 76.08 & 93.86 & -  \\
        \rowcolor{Gray!25}\textbf{CC-Ours} & 80.01 & 93.79 & 77.11 & \textbf{94.21} & 0.44 \\
        \bottomrule
    \end{tabular}}
\end{table*}

\begin{table*}[t!]
    \caption{(\emph{left}) Ablation results on CIFAR100 at \(0.3\) and \(0.5\) IDN. We display accuracy results under different configurations and compare the training time of our approach against DivideMix~\cite{li2020dividemix}. The implementation settings and detailed computational analysis are described in~\cref{subsec:implementation} and \cref{sec:computational_analysis} respectively. (\emph{right above}) Ablation results for initialisation with different \(\epsilon\) values on CIFAR100~\cite{krizhevsky2009learning} at a synthetic noise rate of \(0.5\) IDN~\cite{xia2020part} with DivideMix-Ours. (\emph{right below}) Ablation graph showing the learning pattern of the noise rate parameter estimation under different initialisation values, i.e., \(\epsilon = 0.001\) (blue), \(\epsilon = 0.1\) (orange), \(\epsilon = 0.9\) (red), when trained on DivideMix-ours case at \(0.5\) IDN~\cite{xia2020part} on CIFAR100~\cite{krizhevsky2009learning}. }
    \label{tab:ablation_cifar}
    \begin{minipage}[t]{0.5\linewidth}
        \centering
        \scalebox{0.65}{
            \begin{tabular}{lcc}
            \toprule
            \multirow{2}{*}{\bfseries Models} & \multicolumn{2}{c}{\bfseries Accuracy (\%)} \\
            \cmidrule(lr){2-3}
            & \bfseries 0.3 & \bfseries 0.5 \\
            \midrule
            DivideMix~\cite{li2020dividemix} & 76.33 & 58.61 \\
            DivideMix with fixed \(\epsilon\) & 78.14 & 64.44 \\
            Our method with pre-trained DivideMix & 75.01 & 52.31 \\
            Our method with original DivideMix (no \(\epsilon\)) & 76.20 & 56.30 \\
            \rowcolor{gray!25} \textbf{DivideMix-Ours} & 77.21 & 64.02 \\
            \midrule
            \multicolumn{3}{l}{\bfseries Training Time (GPU-hours)} \\
            \midrule
            DivideMix~\cite{li2020dividemix} & \multicolumn{2}{c}{7.20} \\
            \rowcolor{gray!25} \textbf{DivideMix-Ours} & \multicolumn{2}{c}{8.50}\\
            \rowcolor{gray!25} \textbf{DivideMix-Ours (Mixed Precision)} & \multicolumn{2}{c}{6.10}\\
            \bottomrule
        \end{tabular}
        }
    \end{minipage}%
    \begin{minipage}[c]{0.45\linewidth}
        \centering
        \scalebox{0.5}{
           \begin{tabular}{l c  c}
                \toprule
                \multirow{2}{*}{\bfseries Initialisation} & \bfseries Accuracy  & \bfseries Estimated \\ \\
                \bfseries \(\boldsymbol{\epsilon}\) & (\%) &  \bfseries (\(\boldsymbol{\epsilon}\)) \\
                \midrule
                \(0.001\) & 63.87 & 0.52 \\
                \(0.1\) & 64.01 & 0.53 \\
                \(0.9\) & 63.79 & 0.52 \\
                \bottomrule
                \\
            \end{tabular}
        }
        \centering
        \scalebox{0.75}{
        \begin{tikzpicture}
    \pgfplotstableread[col sep=space, header=true]{images/plots/001.tex} \co
    \pgfplotstableread[col sep=space, header=true]{images/plots/01.tex} \cot
    \pgfplotstableread[col sep=space, header=true]{images/plots/09.tex} \se
    \begin{axis}[
        height = 0.5\linewidth,
        width = 0.75\linewidth,
        xlabel={\textnumero~of epochs T},
        xlabel style={font=\scriptsize},
        xticklabel style = {font=\scriptsize},
        xmin=30,
        xmax=310,
        ylabel={Noise rate \(\epsilon\)},
        ylabel style={font=\scriptsize, yshift=-0.5em},
        yticklabel style = {font=\scriptsize},
        ymin=0,
        legend style={draw=none, font=\tiny, legend cell align={left}, legend columns=1, yshift=0.15em, xshift=0.2em},
        legend image post style={scale=0.45},
        scale only axis
    ]
    \addplot[mark=none, MidnightBlue, thick, dashed] table[x={epochs}, y={e}]{\co};
    \addlegendentry{\(\mathrm{init}(\epsilon) = 0.001\)}
    
    \addplot[mark=none, BurntOrange, thick, dashdotted] table[x={epochs}, y={e}]{\cot};
    \addlegendentry{\(\mathrm{init}(\epsilon) = 0.1\)}

    \addplot[mark=none, BrickRed, solid, thick, solid] table[x={epochs}, y={e}]{\se};
    \addplot[mark=none, style={thin, densely dashed}] coordinates {(30, 0.5) (311, 0.5)};
    \node[above, color=Black, rotate=0] at (90, 0.5) {\scriptsize{ideal ratio}};
    
    \end{axis}
    \node[below right, yshift=7mm, xshift=-22mm] at (current axis.south east) {
        \begin{tikzpicture}
            \draw[BrickRed, solid, thick] (0,0) -- (0.25,0) node[right, black, font=\tiny] { \(\mathrm{init}(\epsilon) = 0.9\)};
        \end{tikzpicture}
    };
\end{tikzpicture}}
    \end{minipage}
\end{table*}

\subsubsection{Real-World Noise}\label{subsubsec:real_world}

We also evaluate our proposed method on various real-world noisy settings regarding test accuracy and estimated noise rates \(\epsilon\) in~\cref{table:redmini,table:clothing1M,table:miniWebvision}. Similarly to the synthetic IDN in~\cref{table:cifar100}, the results show that existing noisy-label robust methods can be easily integrated with our model to outperform current SOTA results for real-world noisy-label datasets. \cref{table:redmini} shows the results on red mini-ImageNet using two configurations, including cases without self-supervision (top part of the table) and with self-supervision (bottom part of the table). The self-supervision DINO pre-training~\cite{caron2021emerging} relies only on images from red mini-ImageNet to enable a fair comparison with existing baselines~\cite{cordeiro2021propmix, garg2023instance}.
Results from~\cref{table:redmini} demonstrate that our approach improves the performance of SOTA methods by a considerable margin in all cases. 
In fact, using estimated noise rate \(\epsilon\) while training InstanceGM~\cite{garg2023instance} without self-supervision shows better performance than existing self-supervised baselines at \(0.4\) noise rate. Moreover, DivideMix~\cite{li2020dividemix}, C2D~\cite{zheltonozhskii2022contrast}, SSR~\cite{feng2021ssr}, and CC~\cite{zhao2022centrality} are used as a baselines for Clothing1M~\cite{xiao2015learning} as shown in~\cref{table:clothing1M}, with results showing slight improvements with the use of our method. 
For mini-WebVision~\cite{li2020dividemix} (validation on ImageNet~\cite{krizhevsky2012imagenet}),
we use
DivideMix~\cite{li2020dividemix}, C2D~\cite{zheltonozhskii2022contrast}, SSR~\cite{feng2021ssr}, and CC~\cite{zhao2022centrality} as baselines (\cref{table:miniWebvision}).
It is worth noting that our results are better than the SOTA for all settings in~\cref{table:miniWebvision}. 

\subsection{Ablation}\label{subsec:ablation}

We show an ablation study with testing accuracy \emph{(top)} and training time \emph{(bottom)} of our approach in~\cref{tab:ablation_cifar} \emph{(left)} on CIFAR100~\cite{krizhevsky2009learning} at \(0.3\) and \(0.5\) IDN~\cite{xia2020part} using  DivideMix~\cite{li2020dividemix} as baseline. 
Detailed complexity analysis is shown in the supplementary material,~\cref{tab:comp_analysis}.
Initially, the accuracy result of baseline DivideMix~\cite{li2020dividemix} is \(76.33\%, 58.61 \%\) for  \(0.3\) and \(0.5\) IDN, respectively.  In the second row, we fix the noise rate \(\epsilon\) at  \(0.3\) and \(0.5\) for DivideMix's sample selection, as explained in~\cref{subsec:noise_rate_integration} (without updating \(\epsilon\)), then the results improved to \(78.14 \%\) and \(64.44 \%\), which is the ideal case (i.e., a perfect noise rate estimation) that motivated our work. In the third case, we use the proposed graphical model with pre-trained DivideMix~\cite{li2020dividemix}  that shows accuracy of \(75.01\%\) and \(52.31\%\). In the next case, the proposed graphical model is trained together with DivideMix~\cite{li2020dividemix} without considering the estimated noise rate \(\epsilon\) for sample selection, which results in accuracy of \(76.20\%\) and \(56.30\%\). In the last row, we show the training of the proposed model with DivideMix~\cite{li2020dividemix}, together with the estimation of noise rate \(\epsilon\), and the selection of samples based on that,  with \(\approx 1\%\) and \(\approx 8\%\) accuracy improvement fo \(0.3\) and \(0.5\) IDN, which is quite close to our ideal case (second row).
We also study the effect that the initialisation of the noise rate parameter \(\epsilon\) plays in our method. As described in~\cref{subsec:implementation}, we randomly initialise \(\epsilon\) within the range (\(0,1\)). To test the robustness of our method to this initialisation, we perform an ablation study by initialising \(\epsilon\) with three different values, namely: \(0.001, 0.1,\) and \(0.9\) on CIFAR100~\cite{kye2022learning} at \(0.5\) IDN~\cite{xia2020part} using the DivideMix-ours approach. Our results, presented in~\cref{tab:ablation_cifar} \emph{(right above and below)}, demonstrate that the initial value of \(\epsilon\) has a limited effect on the model's performance, highlighting the stability of our method across different initialisation of \(\epsilon\).
In the supplementary material, \cref{tab:ssr_supp} \emph{(left)} presents comparative analyses of our model against PartT~\cite{xia2020part} and SSR~\cite{feng2021ssr} on CIFAR100~\cite{krizhevsky2009learning} at \(0.5\) IDN~\cite{xia2020part}, while \cref{tab:ssr_supp} \emph{(right)} showcases similar comparisons with baselines DivideMix~\cite{li2020dividemix} and InstanceGM~\cite{garg2023instance} at a higher noise rate of \(0.6\) IDN~\cite{xia2020part}. Additionally, \cref{table:redmini_low_rate_supp} provides further insights into the performance of DivideMix~\cite{li2020dividemix} on red mini-ImageNet~\cite{jiang2020beyond} at lower noise rates of \(0.1, 0.2\) and \(0.3\). \cref{tab:standard_deviation_supp} illustrates noise rate estimation with standard deviation for our model integrated with DivideMix~\cite{li2020dividemix} and InstanceGM~\cite{garg2023instance} on CIFAR100~\cite{krizhevsky2009learning} under IDN~\cite{xia2020part} settings at \(0.2\), \(0.3\), \(0.4\), and \(0.5\).

\section{Discussion and Conclusion}\label{sec:conclusion}

In this work, we proposed an innovative graphical model for IDN noisy-label learning, focusing on the estimation of the label noise rate that is leveraged to introduce a robust sample selection curriculum. 
Given that SOTA IDN noisy-label learning approaches tend to rely on sample selection methods, our method can be seamlessly integrated to them to improve their performance in many synthetic and real-world benchmarks, including
CIFAR100~\cite{krizhevsky2009learning}, red mini-ImageNet~\cite{jiang2020beyond}, Clothing1M~\cite{xiao2015learning}, mini-WebVision~\cite{li2020dividemix}, and ImageNet~\cite{krizhevsky2012imagenet}. 

From a societal perspective, our methodology offers a positive impact by mitigating biases inherent in noisy data, thereby promoting more fair and accurate machine learning outcomes. It also opens the door to future research avenues, including exploring advanced noise rate estimation methodologies and investigating noise-robust loss functions such as GCE~\cite{zhang2018generalized} and ELR~\cite{liu2020early} with it. 
Our choice of using the same network architecture for clean and noisy labels enables compatibility with co-teaching techniques~\cite{han2018co,li2020dividemix}. However, investigating different network structures for clean and noisy labels remains a potential avenue for future research. Moreover, we aim to delve into its dynamics, uncovering substantial improvements for certain models while observing more marginal enhancements for others.
Our approach also hints at potential breakthroughs in streamlining data annotation challenges. 

\clearpage  

%
%
\bibliographystyle{splncs04}
\bibliography{egbib}

\clearpage
\setcounter{page}{1}
\crefalias{section}{appendix}
\begin{appendices}
\section{Extended Implementation Details}\label{sec:implementation_supp}
In our proposed graphical model, as illustrated in~\cref{fig:graphical_model}, the term \(p(y|x;\theta_{y})\) utilises the baseline classifier. Therefore, the architecture of this classifier remains in alignment with the integrated baseline corresponding to this term.
For the terms \(q(y|x,\hat{y})\) and \(p(\hat{y}|x,y;\theta_{\hat{y}},\epsilon)\), we utilise a network architecture analogous to the baseline classifier present in the state-of-the-art integrated methodologies.
In our methodology, we utilised PyTorch's auto-cast feature to optimise computational efficiency (\cref{tab:comp_analysis}). The training uses stochastic gradient descent (SGD) with a momentum of \(0.9\). The classifiers' learning rate is kept the same as their baseline model. The noise rate \(\epsilon\) is learned using the learnable parameter of the sigmoid activation function, where training uses SGD with a learning rate of \(0.001\) and momentum of \(0.9\). The WarmUp stage also follows the baselines with the following number of epochs: \(30\) for CIFAR100~\cite{krizhevsky2009learning} and red mini-ImageNet~\cite{jiang2020beyond}, \(1\) for Clothing1M~\cite{xiao2015learning}, and \(5\) for mini-WebVision~\cite{li2020dividemix}. The batch sizes used are \(64\) for CIFAR100~\cite{krizhevsky2009learning} and red mini-ImageNet~\cite{jiang2020beyond}, \(32\) for Clothing1M~\cite{xiao2015learning} and mini-WebVision~\cite{li2020dividemix}. Additionally, for the self-supervision variant of red mini-ImageNet~\cite{jiang2020beyond}, we use DINO~\cite{caron2021emerging}, where all the settings of DINO~\cite{caron2021emerging} are unchanged from its original work. DINO is trained on red mini-ImageNet~\cite{jiang2020beyond} and the WarmUp stage is reduced to \(10\) epochs. For Clothing1M~\cite{xiao2015learning}, pre-trained ResNet-50 is used for DivideMix~\cite{li2020dividemix} and CC~\cite{zhao2022centrality}, and  \say{clean data is not used while training}. Similarly, the variant without class balance is used for SSR~\cite{feng2021ssr} on Clothing1M~\cite{xiao2015learning}, and no pre-trained network is used. Moreover, while training C2D~\cite{zheltonozhskii2022contrast} on mini-WebVision~\cite{li2020dividemix}, we use the provided pre-trained classifier ResNet-50 with SimCLR~\cite{khosla2020supervised} for self-supervision.

\section{Empirical Analysis of our Approach}\label{sec:empirical_analysis}

In this section, we compare our sample selection approach with the sample selection methods in DivideMix~\cite{li2020dividemix} (\cref{fig:supp_empirical_graph}) and SSR~\cite{feng2021ssr} (\cref{fig:supp_empirical_graph_ssr}) on CIFAR100~\cite{krizhevsky2009learning} at \(0.5\) IDN~\cite{xia2020part}. We show the F1 score, precision, and ratio of clean samples classified on the last \(50\) training epochs.
\paragraph{F1-Score:} \cref{fig:f1} shows DivideMix's baseline small loss approach~\cite{li2020dividemix} resulting in \(\approx 0.70\), whilst our sample selection approach integrated with DivideMix~\cite{li2020dividemix} reaches around \(0.92\). Our approach integrated with SSR~\cite{feng2021ssr} is shown to achieve around \(0.93\) whereas the baseline SSR shows \(0.94\) in~\cref{fig:f1_ssr}. 
\paragraph{Precision:} \cref{fig:precision} shows the precision comparison, where DivideMix's small loss~\cite{li2020dividemix} reaches around \(0.65\), whereas our approach with DivideMix produces a result around \(0.95\). Additionally,~\cref{fig:precision_ssr} shows that our approach with SSR has a precision of \(0.97\) that is slightly larger than the SSR's  precision of around \(0.95\). 
\begin{figure*}[t!]
    \centering
    \begin{subfigure}[b]{0.3\textwidth}
        \centering
        \begin{tikzpicture}
            \pgfplotstableread[col sep=space, header=true]{csvs/f1_dm.tex} \dmTable
            \pgfplotstableread[col sep=space, header=true]{csvs/f1_OursDM.tex} \myTable

            \begin{axis}[
                height = 0.6\linewidth,
                width = 0.8\linewidth,
                xlabel={\textnumero~of epochs},
                xlabel style={font=\tiny},
                xticklabel style = {font=\tiny},
                xmin=250,
                xmax=301,
                ylabel={F1-score},
                ylabel style={font=\tiny, yshift=-0.5em},
                yticklabel style = {font=\tiny},
                scale only axis
            ]
                \addplot[mark=none, MidnightBlue, thick, solid] table[x={epochs}, y={F1}]{\dmTable};
                \addplot[mark=none, BurntOrange, thick, solid] table[x={epochs}, y={F1}]{\myTable};
            \end{axis}
        \end{tikzpicture}
        \caption{F1 Score}
        \label{fig:f1}
    \end{subfigure}
    \hfill
    \begin{subfigure}[b]{0.3\textwidth}
        \centering
        \begin{tikzpicture}
            \pgfplotstableread[col sep=space, header=true]{csvs/precision_dm.tex} \dmTable
            \pgfplotstableread[col sep=space, header=true]{csvs/precision_OursDM.tex} \myTable

            \begin{axis}[
                height = 0.6\linewidth,
                width = 0.8\linewidth,
                xlabel={\textnumero~of epochs},
                xlabel style={font=\tiny},
                xticklabel style = {font=\tiny},
                xmin=250,
                xmax=301,
                ylabel={Precision},
                ylabel style={font=\tiny, yshift=-0.5em},
                yticklabel style = {font=\tiny},
                legend entries={Small loss, Ours},
                legend style={
                    draw=none, 
                    font=\tiny,
                    at={(0.95,0.5)}, 
                    anchor=south east,
                    cells={anchor=west},
                    legend columns=2, 
                },
                legend image post style={scale=0.5},
                scale only axis
            ]
                \addplot[mark=none, MidnightBlue, thick, solid] table[x={epochs}, y={Precision}]{\dmTable};
                \addplot[mark=none, BurntOrange, thick, solid] table[x={epochs}, y={Precision}]{\myTable};
            \end{axis}
        \end{tikzpicture}
        \caption{Precision}
        \label{fig:precision}
    \end{subfigure}
    \hfill
    \begin{subfigure}[b]{0.3\textwidth}
        \centering
        \begin{tikzpicture}
            \pgfplotstableread[col sep=space, header=true]{csvs/ratio_dm.tex} \dmTable
            \pgfplotstableread[col sep=space, header=true]{csvs/ratio_OursDM.tex} \myTable

            \begin{axis}[
                height = 0.6\linewidth,
                width = 0.8\linewidth,
                xlabel={\textnumero~of epochs},
                xlabel style={font=\tiny},
                xticklabel style = {font=\tiny},
                xmin=250,
                xmax=301,
                ylabel={Ratio of clean data},
                ylabel style={font=\tiny, yshift=-0.5em},
                yticklabel style = {font=\tiny},
                scale only axis
            ]
                \addplot[mark=none, MidnightBlue, thick, solid] table[x={epochs}, y={Ratio}]{\dmTable};
                \addplot[mark=none, BurntOrange, thick, solid] table[x={epochs}, y={Ratio}]{\myTable};

                \addplot[mark=none, style={densely dashed}] coordinates {
                    (250, 0.5)
                    (300, 0.5)
                };
                \node[above, color=Black, rotate=0, ] at (265, 0.5) {\scriptsize{ideal ratio}};
            \end{axis}
        \end{tikzpicture}
        \caption{Clean Classified Ratio}
        \label{fig:ratio}
    \end{subfigure}
    \caption{The graphs above show (\subref{fig:f1})~F1-score, (\subref{fig:precision})~precision, and (\subref{fig:ratio})~ratio of data classified as clean by the sample selection strategy as a function of the last 50 epochs for our approach with DivideMix~\cite{li2020dividemix} (orange) and DivideMix's approach~\cite{li2020dividemix} based on small loss  (blue) on CIFAR-100 \cite{krizhevsky2009learning} with \(0.5\) IDN~\cite{xia2020part} noise rate.}
        \label{fig:supp_empirical_graph}
\end{figure*}
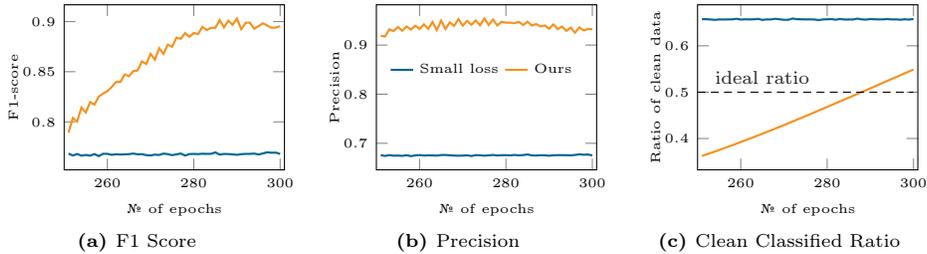
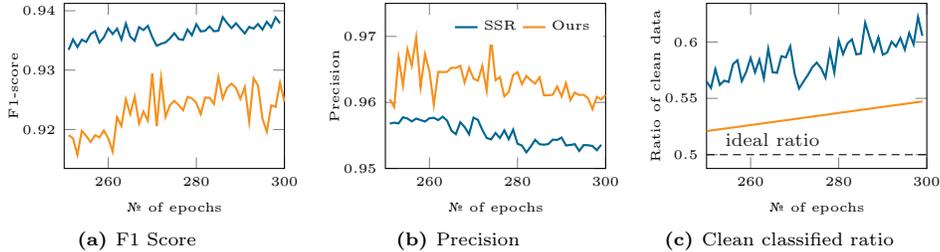
\begin{figure*}[t!]
    \centering
    \begin{subfigure}[b]{0.3\textwidth}
        \centering
        \begin{tikzpicture}
            \pgfplotstableread[col sep=space, header=true]{csvs/f1_ssr.tex} \dmTable
            \pgfplotstableread[col sep=space, header=true]{csvs/f1_OursSSR.tex} \myTable
            \begin{axis}[
                height = 0.6\linewidth,
                width = 0.8\linewidth,
                xlabel={\textnumero~of epochs},
                xlabel style={font=\tiny},
                xticklabel style = {font=\tiny},
                ylabel={F1-score},
                ylabel style={font=\tiny, yshift=-0.5em},
                yticklabel style = {font=\tiny},
                xmin=250,
                xmax=300,
                scale only axis
            ]
                \addplot[mark=none, MidnightBlue, thick, solid] table[x={epochs}, y={F1}]{\dmTable};
                \addplot[mark=none, BurntOrange, thick, solid] table[x={epochs}, y={F1}]{\myTable};
            \end{axis}
        \end{tikzpicture}
        \caption{F1 Score}
        \label{fig:f1_ssr}
    \end{subfigure}
    \hfill
    \begin{subfigure}[b]{0.3\textwidth}
        \centering
        \begin{tikzpicture}
            \pgfplotstableread[col sep=space, header=true]{csvs/precision_ssr.tex} \dmTable
            \pgfplotstableread[col sep=space, header=true]{csvs/precision_OursSSR.tex} \myTable
            \begin{axis}[
                height = 0.6\linewidth,
                width = 0.8\linewidth,
                xlabel={\textnumero~of epochs},
                xlabel style={font=\tiny},
                xticklabel style = {font=\tiny},
                xmin=250,
                xmax=300,
                ymin=0.95,
                ymax=0.975,
                ylabel={Precision},
                ylabel style={font=\tiny, yshift=-0.5em},
                yticklabel style = {font=\tiny},
                legend entries={SSR, Ours}, 
                legend style={
                    draw=none, 
                    font=\tiny,
                    at={(0.99,0.75)}, 
                    anchor=south east,
                    cells={anchor=west},
                    legend columns=2, 
                },
                legend image post style={scale=0.5},
                scale only axis
            ]
                \addplot[mark=none, MidnightBlue, thick, solid] table[x={epochs}, y={Precision}]{\dmTable};
                \addplot[mark=none, BurntOrange, thick, solid] table[x={epochs}, y={Precision}]{\myTable};
            \end{axis}
        \end{tikzpicture}
        \caption{Precision}
        \label{fig:precision_ssr}
    \end{subfigure}
    \hfill
    \begin{subfigure}[b]{0.3\textwidth}
        \centering
        \begin{tikzpicture}
            \pgfplotstableread[col sep=space, header=true]{csvs/ratio_ssr.tex} \dmTable
            \pgfplotstableread[col sep=space, header=true]{csvs/ratio_OursSSR.tex} \myTable
            \begin{axis}[
                height = 0.6\linewidth,
                width = 0.8\linewidth,
                xmin=250,
                xmax=300,
                xlabel={\textnumero~of epochs},
                xlabel style={font=\tiny},
                xticklabel style = {font=\tiny},
                ylabel={Ratio of clean data},
                ylabel style={font=\tiny, yshift=-0.5em},
                yticklabel style = {font=\tiny},
                scale only axis
            ]
                \addplot[mark=none, MidnightBlue, thick, solid] table[x={epochs}, y={Ratio}]{\dmTable};
                \addplot[mark=none, BurntOrange, thick, solid] table[x={epochs}, y={Ratio}]{\myTable};

                \addplot[mark=none, style={densely dashed}] coordinates {
                    (250, 0.5)
                    (300, 0.5)
                };
                \node[above, color=Black, rotate=0, ] at (265, 0.5) {\scriptsize{ideal ratio}};
            \end{axis}
        \end{tikzpicture}
        \caption{Clean classified ratio}
        \label{fig:ratio_ssr}
    \end{subfigure}
    \caption{The graphs above show (\subref{fig:f1})~F1-score, (\subref{fig:precision})~precision, and (\subref{fig:ratio})~ratio of data classified as clean by the sample selection strategy  as a function of the last 50 epochs for our approach with SSR~\cite{feng2021ssr} (orange) and SSR~\cite{feng2021ssr} alone (blue) on CIFAR-100 \cite{krizhevsky2009learning} with \(0.5\) IDN~\cite{xia2020part} noise rate.}
        \label{fig:supp_empirical_graph_ssr}
\end{figure*}

\paragraph{Ratio of samples classified as clean:} \cref{fig:ratio} exhibits the proportion of instances identified as clean. This setting employs a noise rate of \(0.5\). Consequently, in an optimal scenario, the sample selection should classify half of the training samples as clean (ideal case is \(0.5\)).
Our approach with DivideMix~\cite{li2020dividemix} classifies \(0.53\) of the training data as clean, in comparison to DivideMix's~\cite{li2020dividemix} results of around \(0.65-0.70\). Also, our approach with SSR~\cite{feng2021ssr} estimates \(0.53\) (\cref{fig:ratio_ssr}), which is closer to the ideal rate of \(0.5\) than the baseline SSR~\cite{feng2021ssr} which estimates \(0.65-0.70\). 

\section{Additional Experiments and Discussion}\label{sec:further_exp_discussion_supp}

\cref{tab:cifar_iin_supp} shows the outcomes of integrating our model with DivideMix~\cite{li2020dividemix} on IIN~\cite{han2018co} benchmarks under symmetric noise settings for CIFAR10~\cite{krizhevsky2009learning} and CIFAR100~\cite{krizhevsky2009learning} datasets at noise rates of \(0.3, 0.5\) and \(0.7\). 
\cref{tab:ssr_supp} \emph{(left)} shows the results of SSR~\cite{feng2021ssr} and PartT~\cite{xia2020part} on CIFAR100~\cite{krizhevsky2009learning} at \(0.5\) IDN~\cite{xia2020part}, and \cref{tab:ssr_supp} \emph{(right)} shows the results of DivideMix~\cite{li2020dividemix} and InstanceGM~\cite{garg2023instance} on CIFAR100~\cite{krizhevsky2009learning} at \(0.6\) IDN~\cite{xia2020part}. The results are locally reproduced. Estimating noise rate and integrating it with baseline models boost their performances. \cref{table:redmini_low_rate_supp} shows the model performance \emph{(left)} and estimated noise rate \emph{(right)}  of our approach with baseline DivideMix~\cite{li2020dividemix} on red mini-ImageNet~\cite{jiang2020beyond} at low noise rates of \(0.1, 0.2\) and \(0.3\).
\cref{tab:standard_deviation_supp} presents the noise rate estimation with standard deviation of our model integrated with DivideMix~\cite{li2020dividemix} and InstanceGM~\cite{garg2023instance} on CIFAR100~\cite{krizhevsky2009learning} under IDN~\cite{xia2020part} settings at \(0.2, 0.3, 0.4\) and \(0.5\).

\paragraph{Discussion (DivideMix vs other models):} In~\cref{table:cifar100}, both DivideMix-Ours and InstanceGM-Ours present improvements of \(1.5\%\) to \(9\%\) for noise rates$\ge0.4$, but larger improvements for DivideMix-Ours suggest that DivideMix is farther from upper-bound result in that benchmark than InstanceGM. In~\cref{table:redmini}, both DivideMix-Ours and InstanceGM-Ours show comparably significant improvements (from \(3\%\) to \(9\%\)), particularly at noise rate $0.4$, but InstanceGM-SS-Ours shows large improvement only for low noise rate of $0.2$. \cref{fig:bar-chart_various} shows significant improvements in DivideMix~\cite{li2020dividemix} compared to F-DivideMix~\cite{kim2021fine}. This is because we focus on the small-loss hypothesis~\cite{han2018co} instead of the small distance to class-specific
eigenvector~\cite{kim2021fine} for sample selection.  Given that DivideMix~\cite{li2020dividemix} relies on the small-loss hypothesis instead of F-DivideMix's~\cite{kim2021fine} eigenvalue-based sample selection, it is natural that our approach works better for DivideMix~\cite{li2020dividemix}. Similar to IDN, we posit that enhancements offered by our method are correlated with the extent of potential improvement available to the model within the benchmark. 
Furthermore, we aim to explore its dynamics, revealing significant improvements for some models, while noting more modest enhancements for others.

\begin{table}[t]
\centering
\caption{\emph{(top)} Test accuracy \% and \emph{(bottom)} final estimated noise rate \(\epsilon\) on CIFAR10~\cite{krizhevsky2009learning} and CIFAR100~\cite{krizhevsky2009learning} under different symmetric IIN~\cite{han2018co}. Here, we integrate DivideMix~\cite{li2020dividemix} into our proposed model.}
\label{tab:cifar_iin_supp}
\scalebox{0.9}{ 
\begin{tabular}{l | c c c | c c c}
\hline
\textbf{Method} & \multicolumn{3}{c|}{\textbf{CIFAR10-IIN}} & \multicolumn{3}{c}{\textbf{CIFAR100-IIN}} \\
\hline
& \textbf{0.3} & \textbf{0.5} & \textbf{0.7} & \textbf{0.3} & \textbf{0.5} & \textbf{0.7} \\
\hline
DivideMix~\cite{li2020dividemix} & 95.1 & 94.6 & 93.7 & 75.2 & 74.6 & 61.3 \\
\rowcolor{Gray!25}\textbf{DivideMix-Ours}  & 95.9 & 95.2 & 94.1 & 78.9 & 77.0 & 63.5 \\ 
\hline
\multicolumn{7}{c}{} \\
\cline{1-1} \cline{2-4} \cline{5-7}
\multicolumn{7}{l}{\textbf{Estimated Noise Rates}} \\
\hline
DivideMix-Ours  & 0.31 & 0.48 & 0.68 & 0.29 & 0.49 & 0.71 \\
\hline
\end{tabular}}
\end{table}

\begin{table}[t]
    \caption{\emph{(left)} Test accuracy (\(\%\)) of baseline SSR~\cite{feng2021ssr}, PartT~\cite{xia2020part}, and our approach with SSR~\cite{feng2021ssr} and PartT~\cite{xia2020part} on CIFAR100~\cite{krizhevsky2009learning} at \(0.5\) IDN~\cite{xia2020part}. \emph{(right)} Test accuracy (\(\%\)) of baseline DivideMix~\cite{li2020dividemix}, InstanceGM~\cite{garg2023instance} and our approach with DivideMix~\cite{li2020dividemix} and InstanceGM~\cite{garg2023instance} on CIFAR100~\cite{krizhevsky2009learning} at high \(0.6\) IDN~\cite{xia2020part}. All the results are locally reproduced by us and we also show our approach's final estimated noise rates in all cases}\label{tab:ssr_supp}
    \begin{subtable}{.45\linewidth}
        \centering
        \scalebox{0.7}{
            \begin{tabular}{l c c}
                \toprule
                \textbf{Method} & \textbf{Test Accuracy} & \textbf{Noise Estimation} \\
                \midrule
                SSR~\cite{feng2021ssr} & 75.8 & \\
                \rowcolor{Gray!25}\textbf{SSR-Ours} & 76.9 & 0.47 \\
                PartT~\cite{xia2020part} & 56.8 &  \\
                \rowcolor{Gray!25}\textbf{PartT-Ours} & 58.2 & 0.52 \\
                \bottomrule
            \end{tabular}
        }
    \end{subtable}%
    \begin{subtable}{.45\linewidth}
        \centering
        \scalebox{0.7}{
            \begin{tabular}{l c c}
                \toprule
                \textbf{Method} & \textbf{Test Accuracy} & \textbf{Noise Estimation} \\
                \midrule
                DivideMix~\cite{li2020dividemix} & 50.12 & \\
                \rowcolor{Gray!25}\textbf{DivideMix-Ours} & 57.20 & 0.57 \\
                InstanceGM~\cite{garg2023instance} & 72.01 &  \\
                \rowcolor{Gray!25}\textbf{InstanceGM-Ours} & 74.62 & 0.58 \\
                \bottomrule
            \end{tabular}
        }
    \end{subtable}
\end{table}

\begin{table}[t]
    \centering
    \caption{\emph{(left)} Test accuracy \% and \emph{(right)} final estimated noise rate \(\epsilon\) for red mini-ImageNet~\cite{jiang2020beyond} for low noise rates \(0.1, 0.2\) and \(0.3\). We integrate DivideMix~\cite{li2020dividemix} with our method.}
    \label{table:redmini_low_rate_supp}
    \begin{subtable}[t]{.5\linewidth}
        \centering
        \scalebox{0.8}{
        \begin{tabular}{l c c c}
            \toprule
            \multirow{2}{*}{\bfseries Method} & \multicolumn{3}{c}{\bfseries Noise rate} \\
            \cmidrule{2-4}
             & \bfseries 0.1 & \bfseries 0.2 & \bfseries 0.3 \\
            \midrule
            DivideMix~\cite{li2020dividemix} & 52.75 & 50.96 & 48.91 \\
            \rowcolor{Gray!25}\textbf{DivideMix-Ours} & \textbf{54.38} & \textbf{52.89} & \textbf{51.67} \\ 
            \bottomrule
        \end{tabular}}
    \end{subtable}
    \hfill
    \begin{subtable}[t]{.45\linewidth}
        \centering
        \scalebox{0.8}{
        \begin{tabular}{l c c c}
            \toprule
            \multicolumn{4}{l}{\bfseries Estimated noise rates} \\
            \midrule
            \multirow{2}{*}{\bfseries Method} & \multicolumn{3}{c}{\bfseries Actual noise rate} \\
            \cmidrule{2-4}
            & \bfseries 0.1 & \bfseries 0.2 & \bfseries 0.3 \\
            \midrule
            DivideMix-Ours & 0.11 & 0.19 & 0.32 \\
            \bottomrule
        \end{tabular}}
    \end{subtable}
\end{table}

\begin{table}[t]
    \centering
    \caption{Noise rate estimation with standard deviation of our model integrated with DivideMix~\cite{li2020dividemix} and InstanceGM~\cite{garg2023instance} on CIFAR100~\cite{krizhevsky2009learning} under IDN~\cite{xia2020part} settings at \(0.2, 0.3, 0.4\) and \(0.5\).}\label{tab:standard_deviation_supp}
    \scalebox{0.8}{
    \begin{tabular}{l c c c c}
        \toprule
        \multicolumn{4}{l}{\bfseries Estimated noise rates \(\pm\)  \bfseries std dev.} \\
        \midrule
        \multirow{2}{*}{\bfseries Method} & \multicolumn{4}{c}{\bfseries Actual noise rate} \\
        \cmidrule{2-5}
        & \bfseries 0.2 & \bfseries 0.3 & \bfseries 0.4 & \bfseries 0.5 \\
        \midrule
        DivideMix-Ours & 0.18 \(\pm\)  0.002 & 0.34 \(\pm\)  0.003 & 0.43 \(\pm\)  0.006 & 0.53 \(\pm\)  0.004 \\
        InstanceGM-Ours & 0.23 \(\pm\)  0.002 & 0.37 \(\pm\)  0.001 & 0.42 \(\pm\)  0.003 & 0.47 \(\pm\)  0.003 \\
        \bottomrule
    \end{tabular}}
\end{table}

\section{Complexity Analysis}\label{sec:computational_analysis}

\begin{table}[th]
    \centering
    \small
    \caption{Computational analysis of the vanilla DivideMix~\cite{li2020dividemix} and the DivideMix with our noise rate integration.}\label{tab:comp_analysis}
    \scalebox{0.8}{
    \begin{tabular}{l c c}
        \toprule
        \textbf{Model} & \textbf{GFLOPs} $\downarrow$ & \textbf{Throughput (img/sec)} $\uparrow$ \\
        \midrule
        DivideMix~\cite{li2020dividemix} & 1.115 & 465.25 \\
        \rowcolor{gray!25} \textbf{DivideMix-ours} & 1.120 & 897.62 \\
        \bottomrule
    \end{tabular}
    }
\end{table}

In this section, we present the complexity of our method shown in \cref{algorithm:Proposed_Algo} and compare it to some common SOTA methods. To simplify the analysis, we omit the warm-up state and analyse the complexity per an epoch. We also define all the notations used and show in \cref{tab:notations} to ease the analysis. Furthermore, our interest is the complexity induced by the learning algorithm, not the dimension of input data. Thus, we omit the number of data dimension to simplify our analysis.

\begin{table}[t]
    \centering
    \caption{The notations used in the complexity analysis}
    \label{tab:notations}
    \begin{tabular}{l c}
        \toprule
        \bfseries Notation & \bfseries Description \\
        \midrule
        \(N\) & number of training samples \\
        \(C\) & the number of classes \\
        \(\abs{\theta}\) & number of parameters \(\theta\) \\
        \(d\) & the number of dimensions of extracted features \\
        \bottomrule
    \end{tabular}
\end{table}

Our method proposed in \cref{algorithm:Proposed_Algo} consists of three main steps. The complexity of each step is analysed as follows:

\paragraph{Sample selection}
    This step consists of three sub-steps:
    \begin{itemize}
        \item Forward pass to calculate the loss on \(N\) samples: \(\order{N \times \abs{\theta_{y}}}\) (this sub-step can be fast-tracked by parallelism, but we use this to simplify the analysis).,
        \item Sort those \(N\) loss values: \(\order{N \ln{N}}\),
        \item Threshold the clean and noisy-label samples using the obtained noise rate: \(\order{N}\).
    \end{itemize}
    Overall, the complexity of this step is: \(\order{N(\abs{\theta_{y}} + \ln{N} + 1)}\).

\paragraph{Expectation step}
    This step is equivalent to training a deep neural network with the loss function defined in \cref{eq:lower_bound}. In short, it consists of two sub-steps: forward and backward. Note that for simplicity, we perform a Monte Carlo approximation by sampling a single sample from the variational distribution \(q(y | x, \hat{y}; \rho)\).

    The complexity of the forward pass can be decomposed into:
    \begin{itemize}
        \item Forward pass for \(\ln p(\hat{y} | x, y; \theta_{\hat{y}}, \epsilon)\): \(\order{N \abs{\theta_{\hat{y}}}}\),
        \item Forward pass for \(\ln p(y | x; \theta_{y})\): \(\order{N \abs{\theta_{y}}}\)
        \item Forward pass for the entropy \(\mathbb{H}(q)\) (one forward pass to calculate \(y\), then entropy): \(\order{N(\abs{\rho} + C)}\)
    \end{itemize}

    The complexity of the backward pass with autodiff is the same as the forward pass. Thus, the overall complexity in this case is: \(\order{2N(\abs{\theta_{\hat{y}}} + \abs{\theta_{y}} + \abs{\rho} + C)}\).

\paragraph{Maximisation step}
    Similar to the E step, the M step also calculate the lower-bound \(Q\) defined in \cref{eq:lower_bound}. Thus, it shares a similar complexity with the E step. When adding constraints into the objective function of the M step as mentioned in \cref{subsec:noise_rate_integration}, the complexity in this M step is added another term, denoted as \(\order{T_{\mathrm{constraint}}}\).

    The overall complexity in this case is: \(\order{2N(\abs{\theta_{\hat{y}}} + \abs{\theta_{y}} + \abs{\rho} + C) + T_{\mathrm{constraint}}}\).

\paragraph{Overall complexity per epoch}
\begin{equation*}
    \order{ N( 5\abs{\theta_{y}} + \ln{N} + 1 + 4 \abs{\theta_{\hat{y}}} + 4 \abs{\rho} + 4C) + T_{\mathrm{constraint}} }.
\end{equation*}

In general, \(N \ll \min(\abs{\theta_{y}}, \abs{\theta_{\hat{y}}}, \abs{\rho})\) and so is \(C\).
Thus, we can simplify the overall complexity as follows:
\begin{equation}
    \boxed{
    \order{ N( 5\abs{\theta_{y}} + 4 \abs{\theta_{\hat{y}}} + 4 \abs{\rho} ) + T_{\mathrm{constraint}} }.
    }
    \label{eq:overall_complexity}
\end{equation}

Hence, if a base label noise algorithm is used (e.g., DivideMix~\cite{li2020dividemix} or FINE~\cite{kim2021fine}), our method proposed in \cref{algorithm:Proposed_Algo} will add up a complexity of \(N( 5\abs{\theta_{y}} + 4 \abs{\theta_{\hat{y}}} + 4 \abs{\rho} )\).

The complexity term \(T_{\mathrm{constraint}}\) depends on the base label noise method used and are presented as follows:

\subsection{DivideMix as Base Method}
    Note that the DivideMix~\cite{li2020dividemix} used in our method does not require to apply Gaussian mixture modelling to cluster loss values. In our case, we rely on \(\epsilon\) -- the label noise rate -- to threshold the loss values as shown in \cref{eq:sample_selection}. Thus, the additional complexity included in the M step is mainly due to MixMatch~\cite{berthelot2019mixmatch}.
    \begin{itemize}
        \item Data augmentation: \(\order{N}\)
        \item Label augmentation (including predicted label or forward pass): \(\order{N (  \abs{\theta_{y}} + 1)}\)
        \item MixMatch:
            \begin{itemize}
                \item mixup: \(\order{2N}\) (assume that the input \(\abs{x}\) is reasonably small compared to the number of model's parameters)
                \item loss on mixup data (forward pass): \(\order{N \abs{\theta_{y}}}\)
            \end{itemize}
        \item Back propagation to train model with autodiff: \(\order{N \abs{\theta_{y}}}\)
    \end{itemize}

    The additional complexity with DivideMix-based approach can then be presented as:
    \begin{equation}
        T_{\mathrm{DivideMix}} = \order{3 N \abs{\theta_{y}}}
    \end{equation}

    If we assume that the models used are similar: \(\abs{\theta_{\hat{y}}} \approx \abs{\theta_{y}} \approx \abs{\rho}\), then substituting into \cref{eq:overall_complexity} results in approximately 5 times higher than DivideMix~\cite{li2020dividemix}. Note that this calculation ignores the parallelism and mixed-precision. In practice, this number would be much smaller. We also adopt an efficient empirical approach where only one gradient update is executed, effectively reducing computational costs.
    Although the complexity is higher, it is essential to note that our approach incorporates several techniques to expedite the training process. Such an increase in complexity is justifiable, considering the enhanced performance our model offers. 

\subsection{FINE as Base Method}
    The FINE-based approach~\cite{kim2021fine} is similar to the DivideMix-based approach, except the step of sample selection where the distance to the eigenvector of the largest eigenvalue is used as a replacement for loss value. In this case, its complexity includes an extra overhead due to eigen decomposition of \(C\) Gram matrices: \(\order{C d^{3}}\).

    In general, the FINE-based approach~\cite{kim2021fine} has an additional overhead of \(\order{C d^{3}}\) compared to the DivideMix-based approach~\cite{li2020dividemix}.

\end{appendices}
\end{document}